# Uncertainty quantification and exploration-exploitation trade-off in humans

*Antonio Candelieri* [1], *Andrea Ponti* [2], *Francesco Archetti* [2]

[1] University of Milano-Bicocca, Department of Economics, Management and Statistics, Italy
[2] University of Milano-Bicocca, Department of Computer Science, Systems and Communication, Italy

**Abstract**. The main objective of this paper is to outline a theoretical framework to analyse how humans' decision-making strategies under uncertainty manage the trade-off between information gathering (*exploration*) and reward seeking (*exploitation*). A key observation, motivating this line of research, is the awareness that *human learners* are amazingly fast and effective at adapting to unfamiliar environments and incorporating upcoming knowledge: this is an intriguing behaviour for cognitive sciences as well as an important challenge for Machine Learning. The target problem considered is *active learning* in a black-box optimization task and more specifically how the exploration/exploitation dilemma can be modelled within Gaussian Process based Bayesian Optimization framework, which is in turn based on uncertainty quantification. The main contribution is to analyse humans' decisions with respect to Pareto rationality where the two objectives are improvement expected and uncertainty quantification. According to this Pareto rationality model, if a decision set contains a Pareto efficient (*dominant*) strategy, a rational decision maker should always select the dominant strategy over its dominated alternatives. The distance from the Pareto frontier determines whether a choice is (Pareto) rational (i.e., lays on the frontier) or is associated to "exasperate" exploration. However, since the uncertainty is one of the two objectives defining the Pareto frontier, we have investigated three different uncertainty quantification measures and selected the one resulting more compliant with the Pareto rationality model proposed. The key result is an analytical framework to characterize how deviations from "rationality" depend on uncertainty quantifications and the evolution of the reward seeking process.

**Keywords**: Active learning, Pareto analysis, uncertainty quantification, human learning, exploration/exploitation dilemma.

## 1. Introduction

### 1.1 Motivation

When a human – as well as an algorithm – is asked to search for a *target* under limited *resources (trials*, time, effort, or money), he/she has to sequentially perform *queries in a decision/action space* and observe the associated *outcomes* or *rewards*. This activity at all levels of behaviour and time scales of decision-making requires dealing with the *exploration-exploitation* dilemma: e*xploitation* means using the *knowledge* collected so far to get closer to the target (i.e., maximizing immediate reward), while *exploration* means investing resources to acquire more knowledge to update one's beliefs and potentially upset the current belief (i.e., maximizing immediate information gain). The dilemma arises because of the need to *make decisions under uncertainty*: decisions allowing for increasing knowledge do not necessarily lead to the greatest immediate reward (Wilson et al., 2020a; Wilson et al., 2014).

The trade-off between explorative and exploitative behaviours characterizes many disciplines (Berger-Tal et al., 2014) and has originated a multidisciplinary framework that applies to humans, animals, and organizations. The analysis of the strategies implemented by humans in dealing with uncertainty has been an actively researched topic (Schulz et al., 2015; Gershman, 2018; Schulz and Gershman, 2019). A key observation, motivating this line of research, is also the awareness that *human learners* are amazingly fast and effective at adapting to unfamiliar environments and incorporating upcoming knowledge: this is an intriguing behaviour for cognitive sciences as well as an important challenge for Machine Learning.

The reference task considered in this paper is the optimization problem:

$$\mathrm{x}^* = \underset{x \in \Omega \subset \Re^d}{\mathrm{argmax}} f(x) \qquad (1)$$

with $f(x)$ is black box, meaning that its analytical form is not given, no derivatives are available and the value of $f(x)$ can be only known pointwise through expensive and noisy evaluations. Finally, $\Omega$ denotes the *search space*, usually box bounded.

We consider sequential optimization to solve (1). At each iteration $n$, the agent/algorithm chooses a location $x^{(n)}$ and the associated function value is observed, possibly perturbed by noise, $y^{(n)} = f(x^{(n)}) + \varepsilon$. The goal is to get close to $x^*$ within a limited number, $N$, of trials. A related goal is to maximize the Average Cumulative Reward (ACR) over the $N$, of trials, that is $ACR^{(N)} = \frac{1}{N}\sum_{i=1}^{N} y^{(i)}$.

Recently, the Bayesian optimization framework (BO) (Shahriari, et al., 2015; Frazier, 2018; Archetti and Candelieri, 2019) has become one of the most efficient method for solving (1), which is a common problem in many application domains ranging from robotics and engineering design to biomedicine and Automated Machine Learning (Archetti and Candelieri, 2019). BO is based on a *probabilistic surrogate model* approximating $f(x)$, usually a Gaussian Process (GP), and an *acquisition function* (aka *infill criterion* or *utility function*) which balances exploration/exploitation to implement sample efficiency.

Moreover, BO is linked to the ongoing discussion in cognitive science as to whether also humans' strategies are sample efficient: (Borji and Itti, 2013; Candelieri et al., 2020) have been arguing, based on empirical evidence, that strategies adopted by humans in solving global optimization problems can be associated to BO. Evidence of this is captured in Figure 1: compared to other global optimization methods, the estimated location of $x^*$ provided by BO is the closest to the humans' ones.

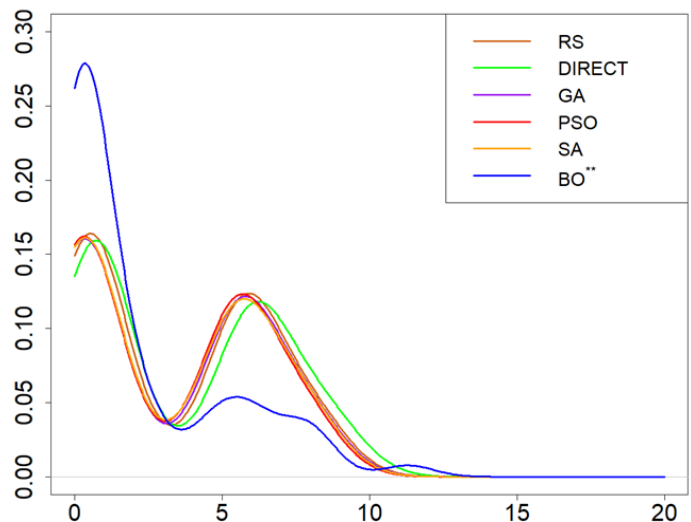

*Figure 1.* Density plot of the distance between the estimated location of $x^*$ provided by humans and those provided by different global optimization strategies: Random Search (RS), DIRECT, Generic Algorithms (GA), Particle Swarm Optimization (PSO), Simulated Annealing (SA) and Bayesian Optimization (from Candelieri et al., 2020).

A caveat is that although the BO is compliant with humans' strategies *over the entire sequential process*, as shown empirically in (Candelieri et al., 2020), it is not necessarily sufficient to capture the exploration-exploitation balance performed by humans *at each decision step*. The working hypothesis of this paper is that this misalignment between Bayesian model of active learning in optimization and humans' strategies might be due to some shortcomings in the general BO's modelling framework: first, the approximation of $f(x)$ depending on decisions and associated outcomes, then the *uncertainty quantification*.

GP modelling and Bayesian learning, first proposed in (Kruschke, 2008; Griffiths et al., 2008) have emerged as central paradigms in modelling human learning, where the GP model is used to approximate the outcome of the next decision conditioned on previous decisions and observed outcomes. Fitting a GP requires to choose, a priori, a *kernel* as covariance function; different kernels are available, each one implying a different characterization for the approximation of $f(x)$. As already stated in (Wilson et al., 2015), it was demonstrated that *"GPs with standard kernels struggle on function extrapolation problems that are trivial for human learners"*. Indeed, they proposed a kernel learning framework to reverse engineer the inductive biases of human learners across a set of behavioural experiments, gaining psychological insights and extrapolating in humanlike ways that go beyond traditional kernels. Different approximations of $f(x)$ can lead to completely different decisions, due to the optimization of the acquisition function whose value depends on the GP model. Many acquisition functions are available; a basic differentiation can be between two "families": the *improvement-based* acquisition functions, searching for the optimum value $f^* = f(x^*)$, and the *information-* *based* acquisition functions, searching for the $x^*$. This distinction is critically important because the two families relate to different quantifications of the uncertainty – as discussed later in the paper – which has been proved to be a key concept also in theories of cognition and emotion (Gershman, 2019). A recent contribution (Bertram et al., 2020) investigates the relationship between *entropy* and the emotional state and the perception of uncertainty.

Uncertainty quantification is not the only modelling issue related to acquisition functions. Recently, some papers proposed to generalize the acquisition mechanism by considering the exploration-exploitation in the framework of the *theory of rational decision making under uncertainty* (Žilinskas and Calvin, 2019). The analysis of the Pareto frontier in the space of the GP's predictive mean and standard deviation offers a set of Pareto-efficient decisions which can be significantly more than those selected through "traditional" acquisition functions. According to Pareto-based rationality model, if a decision set contains a Pareto efficient (*dominant*) strategy, a rational decision maker should always select the dominant strategy over its dominated alternatives. Still, according to a famous Schumpeter quotation (Schumpeter, 1954) traditional decision making under risk *"has a much better claim to being called a logic of choice than a psychology of value"* and indeed deviations from Pareto rational behaviour have been documented in domains like economics, business, but also Reinforcement Learning.

The analysis of violations of dominance in decision-making has become mainstream economics under the name of behavioural economics and prospect theory (Kahneman, 2011): rather than being labelled "irrational", they are just not well described by the rational-agent model. Would a different uncertainty quantification restore rationality? Another key point addressed in this paper is that although BO is the most compliant approach to humans' searching strategies over an entire search task, it could be not sufficiently representative of the exploration-exploitation balance performed by humans at each decision step.

## 1.2 Contributions of this paper

The main contribution of this paper is a methodological framework to analyse how humans' decision-making strategies under uncertainty balance information gathering (*exploration*) and reward seeking (*exploitation*).

The target problem considered in this paper is *active learning* in a black-box optimization task and more specifically how this balancing can be represented by different uncertainty quantifications and exploration/exploitation trade-off in the framework of Gaussian Process modelling.

This required a critical analysis of a large body of results from cognitive science and their relationship with learning and optimization. This has also spawned a more ambitious task: while most of the previous works addressed how people assess the information value of possible queries, in this paper we rather address the issue of the perception of probabilistic uncertainty itself.

This objective has required the development of a software environment for gathering data about human behaviour and analysing them, whose use can be helpful, beyond the specific case, to analyse human strategies in learning problems.

The computational results and their analysis allow to formulate at least a tentative answer to the following research questions:
- Do humans always make "rational" choices (i.e., Pareto optimal decisions between the improvement expected and uncertainty) or, in some cases, they "exasperate" exploration?
- Do different uncertainty quantification measures lead to different classifications of humans' decisions? And which uncertainty quantification measure make humans "more rational"?
- Do deviations from (Pareto) "rationality" and switches towards "exasperated" exploration depend on the evolution of the optimization process measured as the reward collected over the limited number of trials available?

## 1.3 Related works

In Sect. 1.1 we have briefly introduced the issue of uncertainty quantification in humans and its relationship with learning and optimization. Coherently with the centrality of this issue, several research lines have emerged. Here we provide a more specific analysis of the prior work and significant recent results.

An early contribution (Cohen et al., 2007) analyses how humans manage the trade-off between exploration and exploitation in non-stationary environments and links the issue to the Multi Armed Bandit (MAB) problem and Reinforcement Learning. Successively, (Wilson et al., 2014) demonstrates that humans use both *random* and *directed exploration* in a two-arms bandit task. More recently, (Gershman and Uchida, 2019) show how directed exploration in humans amounts to adding an *"uncertainty bonus"* to estimated reward values and how this brings to the *Upper Confidence Bound* acquisition function in MAB (Auer et al., 2002) and BO (Srinivas et al., 2012). The same approach is elaborated in (Schulz and Gershman, 2019), who distinguish between *irreducible uncertainty* related to the reward stochasticity and *uncertainty* which can be reduced through information gathering. In the former the decision strategy is *random search* while for the latter is *directed exploration* which attaches an uncertainty bonus to each decision value. This distinction mirrors the one in Machine Learning between *aleatoric* uncertainty – due to the stochastic variability inherent in querying $f(x)$ – and *epistemic* uncertainty – due to the lack of knowledge about the actual structure of $f(x)$ – which can be reduced by collecting more information. The same point is argued in (Gershman, 2019) which associates *random exploration* to Thompson Sampling, which consists in drawing a sample of $f(x)$ from the GP model and then make the next decision according to the optimization of that sample (Wilson et al., 2020b).

Recent results in the line of research related to brain science are discussed in (Gershman, 2017; Friston et al., 2014). The former analyses the dopamine response in terms of Bayesian Reinforcement Learning, while the second analyses how entropy and expected utility account, respectively, for exploratory and exploitative behaviour, arguing that the dynamics of beliefs updates are consistent with the psychology and anatomy of the dopaminergic system. Moreover, it has been explored how the neuromodulator dopamine plays a central role in encoding and updating of beliefs: *"the level of dopamine is related to the discrepancy between observed and expected reward, known as the reward prediction error (RPE), which serves as a learning signal for updating reward expectations. On the other hand, dopamine also appears to participate in various probabilistic computations, including the encoding of uncertainty and the control of uncertainty-guided exploration"* (Gershman and Uchida, 2019).
These results have been correlated with molecular analysis (Blanco et al., 2015) where it is empirically demonstrated that the carrier of the MET allele in COMT gene will be advantaged in managing the exploration/exploitation dilemma, especially in making choices that maximize long term payoffs.

In the BO research community, recent papers proposed to generalize the acquisition mechanism by considering the exploration-exploitation dilemma as a bi-objective optimization problem: minimizing the predictive mean (associated to exploitation) while maximizing uncertainty, typically the predictive standard deviation (associated to exploration). For instance, in (Žilinskas and Calvin, 2019) the important result is that two well-known acquisition functions, specifically Probability of Improvement (PI) and Expected Improvement (EI), are special cases of this bi-optimization framework, because they lay on the Pareto frontier of all the predictive mean and standard deviation pairs computed for – theoretically – every possible decision. The mean-variance framework has been also considered in (Iwazaki et al., 2020), for multi-task, multi-objective and constrained optimization scenarios. (De Ath et al., 2019; De Ath et al., 2020) show that taking a decision by randomly sampling from the Pareto frontier can outperform other acquisition functions. The main motivation is that the Pareto frontier offers a set of Pareto-efficient decisions wider than that allowed by "traditional" acquisition functions. (Paria et al., 2020) introduce a sampling which can be focused on a specific subregion of the Pareto frontier.

The issue of deviations from Pareto optimality has become a central topic in behavioural economics from the seminal work in (Tversky and Kahneman, 1989) to (Kourouxous and Bauer, 2019) which identifies the most common causes for violations of dominance, namely *framing* (i.e., presentation of a decision problem), *reference points* (i.e., a form of prior expectation), *bounded rationality* and *emotional responses*. Emotions impact decisions by influencing preferences, triggering ad hoc heuristics, or causing attention shifts to catastrophic outcomes.

A recent important contribution is (Sandholtz, 2020) which tackles the problem to infer, given the observed search path generated by a human subject in the execution of a black box optimization task, the unknown acquisition function underlying the sequence. It is to be remarked that this analysis requires restrictive assumptions on the objective function like convexity and smoothness which cannot be assumed in black box problems. For the solution of this problem, referred to as Inverse Bayesian Optimization (IBO), a probabilistic framework for the non-parametric Bayesian inference of the acquisition function is proposed, performed on a set of possible acquisition functions.

This paper is organized as follows. Section 2 introduces the basic definitions about Gaussian Process regression and how different acquisition functions deal with the exploration/exploitation dilemma and use different uncertainty quantification. Section 3 introduces three specific uncertainty quantification measures, the definition of Pareto optimality and the differences induced by different Gaussian Process modelling options (i.e., kernels) and uncertainty quantifications measures. Section 4 introduces the experimental framework used for data collection about the strategies applied by humans and the proposed analytical framework. Section 5 describes the relevant results obtained by the application of the analytical framework. Finally, Section 6 outlines the conclusions about this study and the perspective of future works.

## 2. Materials and methods

### 2.1 Gaussian Process regression

A GP is a *random distribution over functions* $f: \Omega \subset \mathfrak{R}^d \to \mathfrak{R}$ denoted with $f(x) \sim GP(\mu(x), k(x, x'))$ where $\mu(x) = \mathbb{E}(f(x)): \Omega \to \mathfrak{R}$ is the mean function of the GP and $k(x, x'): \Omega \times \Omega \to \mathfrak{R}$ is the *kernel* or *covariance function*. One way to interpret a GP is as a collection of correlated random variables, any finite number of which have a joint Gaussian distribution, so $f(x)$ can be considered as a sample drawn from a multivariate normal distribution. In Machine Learning, GP modelling is largely used for both classification and regression tasks (Williams and Rasmussen, 2006; Gramacy, 2020), providing probabilistic predictions by conditioning $\mu(x)$ and $\sigma^2(x)$ on a set of available data/observations.

Let denote with $X_{1:n} = \{x^{(i)}\}_{i=1,\ldots,n}$ a set of $n$ locations in $\Omega \subset \mathfrak{R}^d$ and with $y_{1:n} = \{f(x^{(i)}) + \varepsilon\}_{i=1,\ldots,n}$ the associated function values, possibly noisy with $\varepsilon$ a zero-mean Gaussian noise $\varepsilon \sim \mathcal{N}(0, \lambda^2)$. Then $\mu(x)$ and $\sigma^2(x)$ are the GP's posterior predictive mean and standard deviation, conditioned on $X_{1:n}$ and $y_{1:n}$ according to the following equations:

$$\mu(x) = k(x, X_{1:n}) [K + \lambda^2 I]^{-1} y_{1:n} \qquad (2)$$

$$\sigma^2(x) = k(x, x) - k(x, X_{1:n}) [K + \lambda^2 I]^{-1} k(X_{1:n}, x) \qquad (3)$$

where $k(x, X_{1:n}) = \{k(x, x^{(i)})\}_{i=1,\ldots,n}$ and $K \in \mathfrak{R}^{n \times n}$ with entries $K_{ij} = k(x^{(i)}, x^{(j)})$.

The choice of the kernel establishes prior assumptions over the structural properties of the underlying (aka latent) function $f(x)$, specifically its smoothness. However, almost every kernel has its own hyperparameters to tune – usually via Maximum Log-likelihood Estimation (MLE) or Maximum A Posteriori (MAP) – for reducing the potential mismatches between prior smoothness assumptions and the observed data. Common kernels for GP regression – considered in this paper – are:

- Squared Exponential: $k_{SE}(x, x') = e^{-\frac{\|x - x'\|^2}{2\ell^2}}$
- Exponential: $k_{EXP}(x, x') = e^{-\frac{\|x - x'\|}{\ell}}$
- Power-exponential: $k_{PE}(x, x') = e^{-\frac{\|x - x'\|^p}{\ell^p}}$
- Matérn3/2: $k_{M3/2}(x, x') = \left(1 + \frac{\sqrt{3}\|x - x'\|}{\ell}\right) e^{-\frac{\sqrt{3}\|x - x'\|}{\ell}}$
- Matérn5/2: $k_{M5/2}(x, x') = \left[1 + \frac{\sqrt{5}\|x - x'\|}{\ell} + \frac{5}{3}\left(\frac{\|x - x'\|}{\ell}\right)^2\right] e^{-\frac{\sqrt{5}\|x - x'\|}{\ell}}$

The main well-known disadvantage of GP modelling is its cubic complexity due to the inversion of the matrix $[K + \lambda^2 I]$.

## 2.2 Dealing with the exploration-exploitation dilemma

Global optimization methods differ one from another in how they generate the next decision (i.e., location) $x^{(n+1)}$. To do this, BO fits a GP according to (2-3) and where $X_{1:n} = \{x^{(i)}\}_{i=1,\ldots,n}$ and $y_{1:n} = \{y^{(i)}\}_{i=1,\ldots,n}$ are the two sequences of, respectively, decisions made and associated observed outcomes. Then, an acquisition function, combining GP's $\mu(x)$ and $\sigma(x)$, is optimized to obtain $x^{(n+1)}$, while dealing with the exploration-exploitation trade-off.

*2.2.1. Improvement-based acquisition functions*

Acquisition functions belonging to this "family" are aimed at searching for $f^* = \max_{x \in \Omega \subset \Re^d} f(x)$ – instead of searching for $x^* = \underset{x \in \Omega \subset \Re^d}{\mathrm{argmax}\,} f(x)$ – and are characterized by "mixing" GP's mean and standard deviation to balance between exploitation and exploration in the choice of $x^{(n+1)}$. Common acquisition functions from this family are Probability of Improvement (PI) (Kushner, 1964), Expected Improvement (EI) (Močkus, 1975) and GP Confidence Bound (i.e., Upper Confidence Bound, UCB, for minimization) (Srinivas et al., 2012):

$$PI(x) = \Phi\left(\frac{\mu(x) - y^+}{\sigma(x)}\right)$$

$$EI(x) = (\mu(x) - y^+) \Phi\left(\frac{\mu(x) - y^+}{\sigma(x)}\right) + \sigma(x) \phi\left(\frac{\mu(x) - y^+}{\sigma(x)}\right)$$

$$UCB(x) = \mu(x) + \sqrt{\beta} \sigma(x)$$

where $\Phi$ and $\phi$ are the standard normal cumulative distribution function (cdf) and the standard normal probability density function (pdf). Since $PI(x)$ and $EI(x)$ are biased to exploration, an additional parameter $\xi$ can be included in the numerator of the arguments of $\Phi$ and $\phi$ to increase exploration (Brochu et al., 2010). Alternatively, an exploration enhanced EI has been recently (Berk et al., 2018) while (Preuss and Von Toussaint, 2018) (deterministically) alternates between maximization of EI and maximization of GP's predictive variance to switch between exploitative and explorative decisions.
GP-UCB, it is also classified as an *optimistic policy*, because it chooses $x^{(i+1)}$ depending on the most optimistic value for $f(x)$ under the current GP. From a cognitive point of view, (Wu et al., 2018) analysed the human search strategy, under a limited number of trials, concluding that GP-UCB offers the best option for modelling the exploitation-exploration trade-off adopted by the humans. Furthermore, contrary to $PI(x)$ and $EI(x)$ – at least to their original formulations – GP-UCB is more flexible, thanks to its own hyperparameter $\beta$, whose value can be set up to give a different relevance to exploitation and exploration in choosing $x^{(n+1)}$ or it can be scheduled to adapt the balance between exploitation and exploration along the optimization process. While it is empirically suggested to apply a decreasing schedule for $\beta$ (i.e., preferring exploration at the beginning and then moving towards exploitation), in (Srinivas et al., 2012) a convergence proof is given for an increasing scheduling of $\beta$, aimed at avoiding to getting stuck at local optima. However, (Berk et al., 2020), has recently obtained better performance by randomly sampling $\beta$ from a given distribution. They proved that this allows to identify more suitable $\beta$ values and to outperform "traditional" GP-CB on a range of synthetic and real-world problems.

*2.2.2 Information-based acquisition functions*

*Information-based* acquisition functions (Hennig and Schuler, 2012) relies on an information-theoretic perspective, that is choosing $x^{(n+1)}$, given $D_{1:n} = (X_{1:n}, y_{1:n})$, to maximize the information about the location of $x^* = \underset{x \in \Omega \subset \Re^d}{\mathrm{argmax}\,} f(x)$.

*Information gain* measures how informative is a set of observations, $D_{1:n} = (X_{1:n}, y_{1:n})$, and it is defined as the mutual information between $y^{(n+1)}$ and $y_{1:n}$:

$$I(y_{1:n}; y^{(n+1)}) = H(y_{1:n}) - H(y_{1:n}|y^{(n+1)})$$

and where $H(p(\alpha)) = -\int p(\alpha) \log p(\alpha) \, d\alpha$ is the differential entropy of a generic distribution $p(\alpha)$ and measures the amount of uncertainty in $p(\alpha)$. In the discrete case, that is related to a discrete random variable A, differential entropy is defined as $H(A) = \sum_{\alpha \in A} p(\alpha) \log \frac{1}{p(\alpha)}$.

Two important acquisition functions from this family are Entropy Search (ES) (Hennig and Schuler, 2012) and Predictive Entropy Search (PES) (Hernandez-Lobato et al., 2014). Both use differential entropy to characterize the uncertainty about the location of the optimizer, $x^*$. More specifically, the aim is to choose the next decision $x$ which maximizes the expected uncertainty reduction:

$$ES(x) = H(p(x^*|D_{1:n})) - \mathbb{E}[H(p(x^*|D_{1:n} \cup \{x, y\}))]$$

$$PES(x) = H(p(y|D_{1:n}, x)) - \mathbb{E}[H(p(y|D_{1:n}, x, x^*))]$$

The main difference is that ES uses the expectation over $p(x^*|D_{1:n})$, while PES uses expectation over $p(y|D_{1:n}, x)$. They are anyway analytically intractable and are approximated via expensive computations which requires to sample a set of paths from the GP posterior, at each BO iteration, and compute their optima to estimate the differential entropy. Moreover, computational cost drastically increases with the dimensionality of the search space. Therefore, ES and PES are useful just in the case that $f(x)$ is extremely expensive to evaluate, so that the cost for sampling from GP can be considered negligible. Due to these limitations, the Max-value Entropy Search (MES) acquisition function has been recently proposed (Wang and Jegelka, 2017), where the uncertainty about $x^*$ is replaced with the uncertainty about $y^*$:

$$MES(x) = I(\{x, y\}; y^*|D_{1:n}) = H(p(y)|D_{1:n}, x) - \mathbb{E}[H(p(y|D_{1:n}, x, y^*))]$$

MES requires to sample $y^*$ (instead of $x^*$) which can be done by sampling from the GP posterior or from a Gumbel distribution, as also proposed in (Wang and Jegelka, 2017). In (Wang et al., 2016) the relation between MES and other popular acquisition functions has been demonstrated, including ES, UCB and PI, which have anyway empirically underperformed MES on several optimization tasks. Although MES is still based on entropy, the estimation of its information gain via sampling is more efficient, because $y^*$ lays in a one-dimensional space.

Linked to sampling from GP posterior is Thompson Sampling (TS), which can be also considered as a sequential optimization strategy *per-se*. Iteratively, TS draws a path by sampling from the GP posterior and then minimize it to obtain $x^{(n+1)}$ as a possible estimation of the location of $x^*$. After $y^{(n+1)}$ is observed, the GP is updated, and TS continues until a termination criterion is met. An analysis on TS has been recently proposed in (Russo & Van Roy 2016), concluding that TS is biased towards exploitation and suggesting that an $\varepsilon$-greedy version of TS can lead to a better performance (i.e., randomly selecting $x^{(n+1)}$ within the search space, with probability $\varepsilon$, or performing TS with probability $1 - \varepsilon$).

An efficient sampling procedure has been recently proposed in (Hahn et al., 2019) (Wilson et al., 2020b). Sampling from GP posterior is at the basis of information-based acquisition functions, described in the following section.

The distinction between the two families of acquisition functions is relevant in terms of computational cost but, more relevant at least in this paper, is their difference in terms of the uncertainty quantification, providing more options for modelling the uncertainty quantification made by a human.

## 2.3 The problem of uncertainty quantification

From the viewpoint of Machine Learning, uncertainty quantification plays a pivotal role in reduction of errors during learning, optimization and decision making. In (Abdar et al., 2020) a wide survey of different uncertainty quantification methods is provided, considering many application fields, such as computer vision (e.g., self-driving cars and object detection), image processing (e.g., image restoration), medical image analysis (e.g., medical image classification and segmentation), natural language processing (e.g., text classification, social media texts and recidivism risk-scoring).

In decision making, uncertainty is usually associated to exploration: when the uncertainty is "larger" than the possible estimated improvement, then it could be more profitable to adopt an explorative behaviour and acquire more knowledge about $f(x)$.

From a modelling perspective, uncertainty can be split in *aleatoric* and *epistemic* (Der Kiureghian and Ditlevsen, 2009; Kendall and Gal, 2017), where the aleatoric uncertainty is randomness proper in the evaluation of $f(x)$ (usually named "noise") and cannot be reduced, while the epistemic uncertainty depends on the model and can be reduced by collecting more data. In many applications it could be interesting to separate the two types of uncertainty: in (Depeweg et al., 2018) two possible decompositions are described, based on two uncertainty quantification metrics that are variance and entropy. This means that information-based and entropy-based acquisition functions use two different metrics to quantify uncertainty, leading to different trade-off between exploitation and exploration, given the same approximation of the objective function.

From an informal – yet more intuitive – point of view, uncertainty about a decision is the amount of lack of knowledge about it, increasing with the "distance" from decisions already performed and where "distance" can be any suitable metric to compare two decisions. When decisions are locations in a search space, as in this paper, any spatial distance can be considered: an example of this uncertainty quantification has been recently proposed in (Bemporad, 2020) which uses Radial Basis Functions (RBF) as surrogate model and an inverse distance weighting such that the proposed distance is zero at sampled points and grows in between. Although GP's predictive standard deviation usually shows a similar behaviour,it exhibit , in some situations , *variance starvation* (Wang et al., 2018), consisting in an underestimation of variance scale compared to mean scale which can significantly reduce exploration chances in some portions of the search space.

Moreover, GP modelling could be also drastically affected by wrong choices about its prior – kernel type, *in primis* – resulting in possible misleading uncertainty quantification and, consequently, suboptimal exploration. This issue has been recently addressed in (Neiswanger and Ramdas, 2020), in which authors do not assume correctness of the GP prior and generate a confidence sequence for $f(x)$ function using martingale techniques. Cognitive theories of emotion define uncertainty as a cognitive component characterizing emotional states. Finally, when humans' decisions are analysed, there is still another relevant lack in mathematical methods for uncertainty quantification as recently demonstrated in (Schultz et al., 2019) and (Bertram et al., 2020), which have investigated the role of emotion in judgment, risk assessment, and decision making under uncertainty and the different kinds of entropy which can be used to quantify uncertainty in the Sharma-Mittal space of entropy measures. Emotional states are significantly connected with subjective uncertainty estimation. While emotions such as anger and pride are associated to low uncertainty, anxiety and curiosity are associated to high uncertainty. There is not – at the authors' knowledge – any mathematical *"trick"* to implement an emotion-related uncertainty quantification in BO (whichever might be the analogous of "emotion" for an algorithm). Indeed, their conclusion is that emotional conditions have no effect on uncertainty appraisal. Sharma-Mittal entropy uses a parametrised family of surprise functions but effect on the entropy parameters driven by the difference between control and emotional conditions.

## 3. Analytical Framework

### 3.1 Definition of uncertainty quantification measures

Let $\mathcal{K}$ denotes the set of kernels to choose as GP's prior. In this study $\mathcal{K} = \{k_{SE}, k_{EXP}, k_{PE}, k_{M3/2}, k_{M5/2}\}$. Let $\zeta(x)$ denotes the improvement expected by querying the objective function at location $x$, depending on the GPs' posterior (i.e., one GP for each kernel in $\mathcal{K}$). Formally, $\zeta(x) = \mu(x) - y^+$, where $y^+ = \max_{i=1,\dots,n}\{y^{(i)}\}$ because we are considering $\max_{x \in \Omega \subset \mathbb{R}^d} f(x)$. Then, let denote with $\mathcal{U}$ the set of possible uncertainty quantification measures. In this paper we consider the following three alternatives:

- GP's predictive standard deviation, namely $\sigma(x)$. Typically adopted as uncertainty measure in the *improvement-based* acquisition functions.

- GP's differential entropy. For a GP it is given by $H(y|X_{1:n}) = \frac{1}{2}\log\det(K) + \frac{d}{2}\log\det(2\pi e)$, where $K \in \Re^{n \times n}$ with entries $K_{ij} = k(x^{(i)}, x^{(j)}), \forall x^{(i)}, x^{(j)} \in X_{1:n}$ (Williams and Rasmussen, 2006). However, the GP's differential entropy does not depend on the location $x$, but it is just a scalar measure of the uncertainty of the overall GP posterior distribution. Thus, we introduce – following this bullet list – a location-dependent measure of entropy, denoted with $h(x)$, as a possible entropy-based location dependent uncertainty quantification. With respect to the entropy-based uncertainty quantification, and starting from the GP's differential entropy formula, we define:

$$h(x) = H(y|\{X_{1:n} \cup \{x\}\}) = \frac{1}{2}\log\det(K') + \frac{d}{2}\log\det(2\pi e) \quad (4)$$

where $K' \in \Re^{(n+1) \times (n+1)}$ with entries $K'_{ij} = k(x^{(i)}, x^{(j)}), \forall x^{(i)}, x^{(j)} \in \{X_{1:n} \cup \{x\}\}$. This allows us to estimate the entropy-based uncertainty at any location $x$ depending on all the previous decisions $X_{1:n}$. It is important to remark that, analogously to $\sigma(x)$, also $h(x)$ depends – indirectly – on both decisions and outcomes through the kernel function, whose hyperparameters are tuned depending on $D_{1:n}$.

- Distance from previous decisions, inspired from (Bemporad, 2020) and denoted by $z(x)$:

$$z(x) = \begin{cases} 0 & \text{if } \exists\, x^{(i)} \in X_{1:n} : \|x - x^{(i)}\|_2^2 = 0 \\ \frac{2}{\pi}\tan^{-1}\left(\frac{1}{\sum_{j=1}^n w_j(x)}\right) & \text{otherwise} \end{cases} \quad (5)$$

with $w_j(x) = \dfrac{e^{-\|x-x^{(j)}\|_2^2}}{\|x-x^{(j)}\|_2^2}$.

Thus, $z(x)$ is zero at sampled points and grows in between; $\tan^{-1}$ is introduced to damp the growth of $z(x)$ when $x$ is located far away from all sampled points. Contrary to $\sigma(x)$ and $h(x)$, which depend also on $y$, the uncertainty quantification measure $z(x)$ depends only on $X_{1:n}$, that is it depends only on how decisions "cover" the search space, irrespectively to their outcomes $y_{1:n}$. Although $z(x)$ intentionally ignores a portion of collected knowledge (i.e., outcomes of decisions), as main advantage it does not suffer, "by design", from variance starvation.

In Figure 2 a simple 1D example is reported to show the differences between the three uncertainty quantification measures, given the same set of previous decisions and GP model. Just for visualization purposes, each uncertainty quantification measure has been scaled in [0,1]. In this specific case, almost all the intervals between two successive decisions are "equally uncertain", according to $\sigma(x)$ and $h(x)$, where "equally uncertain" means that $\sigma(x) \approx \sigma(x')$ – as well as $h(x) \approx h(x')$ – $\forall x, x' \notin X_{1:n}$. On the contrary, the value of $z(x)$ changes over the search space providing a different quantification of uncertainty at every location.

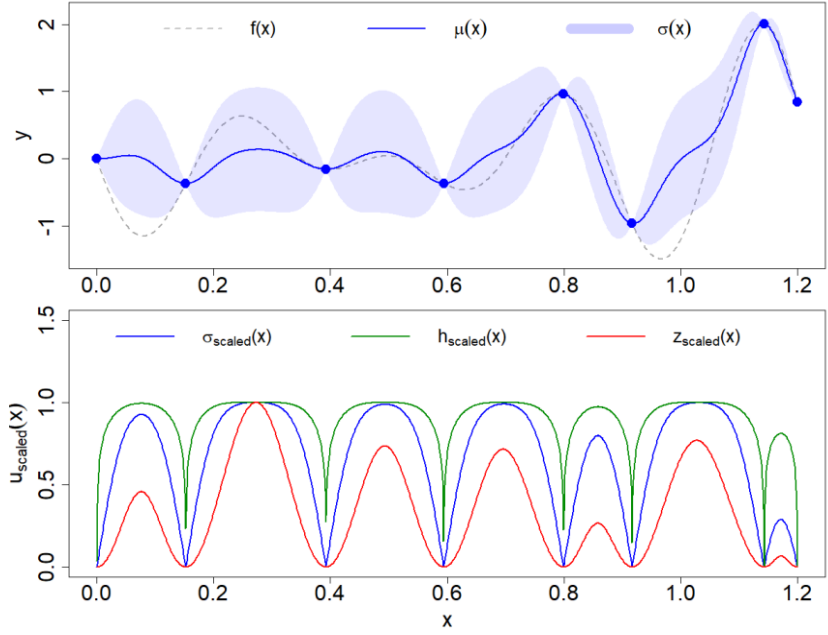

*Figure 2. Differences between the three uncertainty quantification measures considered on a simple 1D example. Top: a black box function $f(x)$, eight observations, the GP's posterior mean and standard deviation. Bottom: the amount of location-dependent uncertainty given by the three uncertainty quantification measures (values are scaled in [0,1] for visualization purposes.*

### 3.2 Pareto rationality

Given the GP conditioned on the decisions performed so far, it is possible to map the next decision $x^{(n+1)} \in \Omega$ – whichever it is – as a bi-objective choice, with objectives $\zeta(x)$ and $u(x) \in \mathcal{U}$ (both to be maximized).
Pareto rationality is the theoretical framework to analyse multi-objective optimization problems where $q$ objective functions $\gamma_1(x), \ldots, \gamma_q(x)$ where $\gamma_i(x): \to \mathbb{R}$ are to be simultaneously optimized in $\Omega \subseteq \mathbb{R}^d$. We use the notation $\boldsymbol{\gamma}(x) = \big(\gamma_1(x), \ldots, \gamma_q(x)\big)$ to refer to the vector of all objectives evaluated at a location $x$. The goal in multi-objective optimization is to identify the Pareto frontier of $\boldsymbol{\gamma}(x)$.
To do this we need an ordering relation in $\mathbb{R}^q$: $\boldsymbol{\gamma} = (y_1, \ldots, y_q) \preccurlyeq \boldsymbol{\gamma}' = (y_1', \ldots, y_q')$ if and only if $\gamma_i \leq \gamma_i'$ for $i = 1, \ldots, q$. This ordering relation induces an order in $\Omega$: $x \preccurlyeq x'$ if and only if $\boldsymbol{\gamma}(x) \preccurlyeq \boldsymbol{\gamma}(x')$.
We also say that $\gamma'$ dominates $\gamma$ (strongly if $\exists\, i = 1, \ldots, q$ for which $\gamma_i < \gamma_i'$). The optimal non-dominated solutions lay on the so-called Pareto frontier.
The interest in finding locations $x$ having the associated $\boldsymbol{\gamma}(x)$ on the Pareto frontier is clear: they represent the trade-off between conflicting objectives and are the only ones, according to the Pareto rationality, to be considered.
In this paper $q = 2$, with $\gamma_1(x)=\zeta(x)$ and $\gamma_2(x) = u(x) \in \mathcal{U}$. Both the objectives are not expensive to evaluate, therefore the Pareto frontier can be easily approximated by considering a fine grid of locations in $\Omega$ without the need to resort to methods approximating expensive Pareto frontiers within a limited number of evaluations, such as in (Zuluaga et al., 2013).
Thus, we approximate our Pareto frontier by sampling a grid of $m$ points in $\Omega$, denoted by $\widehat{\mathbf{X}}_{1:m} = \{x^{(j)}\}_{j=1,\ldots,m}$, and then computing the associated pairs $\Psi_{1:m} = \big\{\big(\zeta(x^{(j)}), u(x^{(j)})\big)\big\}_{j=1,\ldots,m}$.
It is important to remark that a large value of $m$ is needed to have a good approximation of the Pareto frontier but this is not an issue because the computational cost is dominated by conditioning the GP on observations (i.e., $\mathcal{O}(n^3)$, with $n \ll m$) instead of making predictions (i.e., inference). The Pareto frontier can be approximated as:

$$\mathcal{P}(\Psi_{1:m}) = \{\psi \in \Psi_{1:m}: \forall\, \psi' \in \Psi_{1:m}\ \psi \succ \psi'\}$$

where $\psi = \big(\zeta(x), u(x)\big)$ and $\psi' = \big(\zeta(x'), u(x')\big)$, and $\psi \succ \psi' \iff \zeta(x) > \zeta(x') \wedge u(x) > u(x')$.

Figure 3 shows an example of Pareto frontier for $\zeta(x)$ and $u(x) = \sigma(x)$. First five charts, top-left to bottom-right, depict $\Psi_{1:m}$ and the associated $\mathcal{P}(\Psi_{1:m})$ for each kernel in $\mathcal{K} = \{k_{SE}, k_{EXP}, k_{PE}, k_{M3/2}, k_{M5/2}\}$, separately. The last chart (bottom-right) compares only the five Pareto frontiers, better highlighting the role of the GP kernel. For this example, $f(x)$ is the Branin-Hoo (Jekel and Haftka, 2019) function in $\Omega: [-5; 10] \times [0; 15]$, $m = 1976$ (related to a grid $76 \times 26$, obtained by using a step of 0.2 on each dimension).

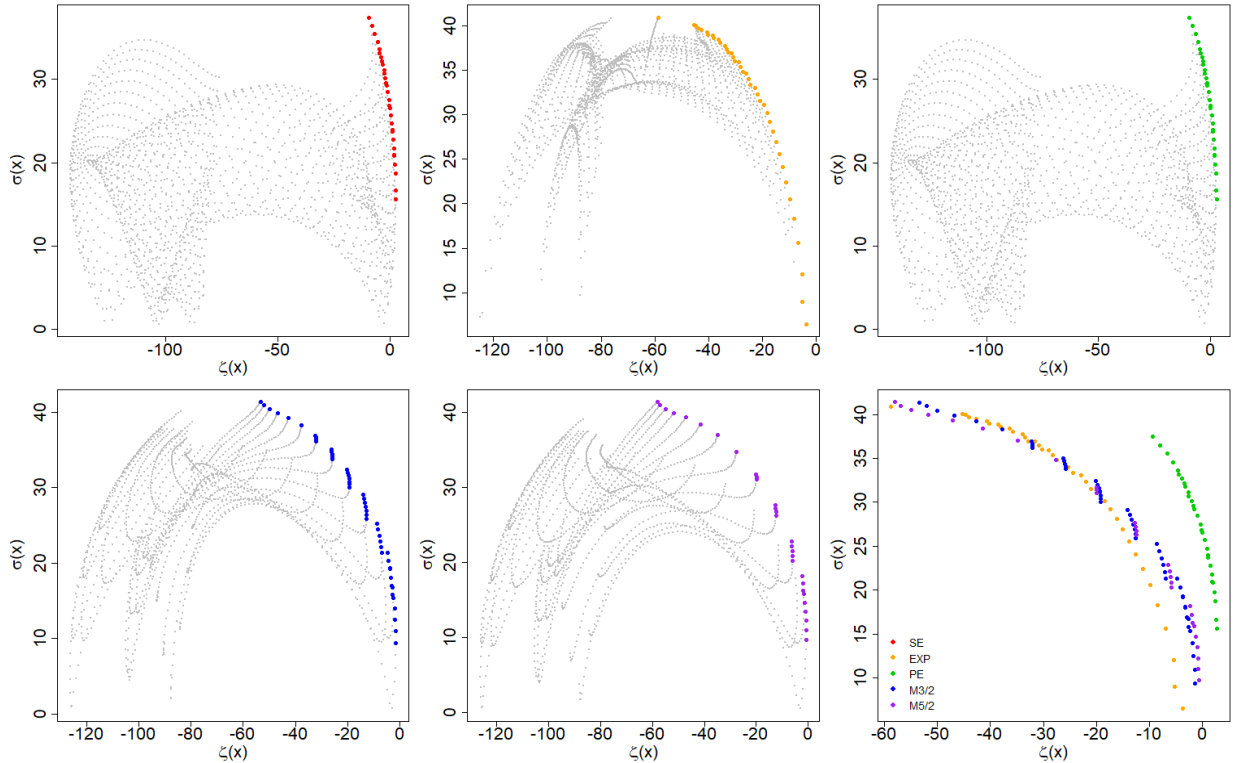

***Figure 3**. Pareto frontiers obtained by using the GP's posterior standard deviation as uncertainty quantification (i.e., $u(x) = \sigma(x)$). Five different kernels are used to fit as many GPs, leading to as many Pareto frontiers. Last chart (bottom-right) depicts the five frontiers all together for an easier comparison.*

Similar charts are reported in Figure 4 and Figure 5 for the other two uncertainty quantification measures.

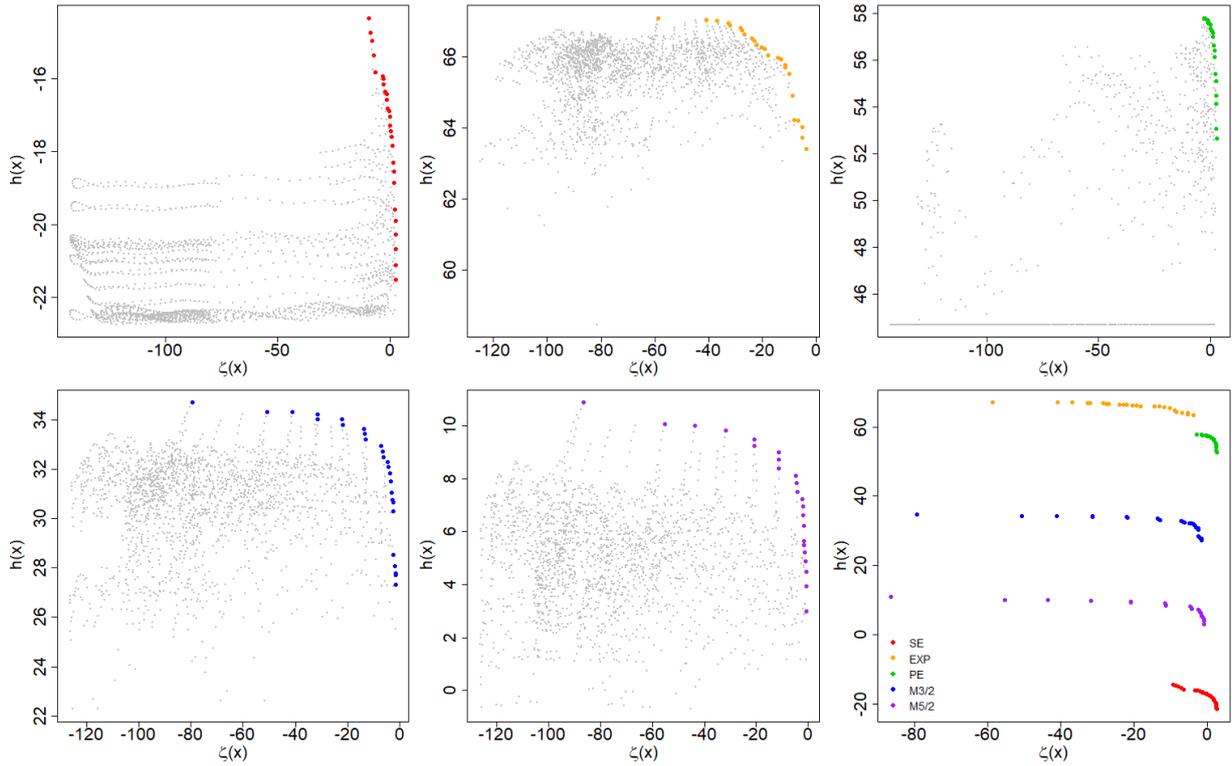

***Figure 4***. *Pareto frontiers obtained by using an entropy-based uncertainty quantification (i.e., $u(x) = h(x)$). Five different kernels are used to fit as many GPs, leading to as many Pareto frontiers. Last chart (bottom-right) depicts the five frontiers all together for an easier comparison.*

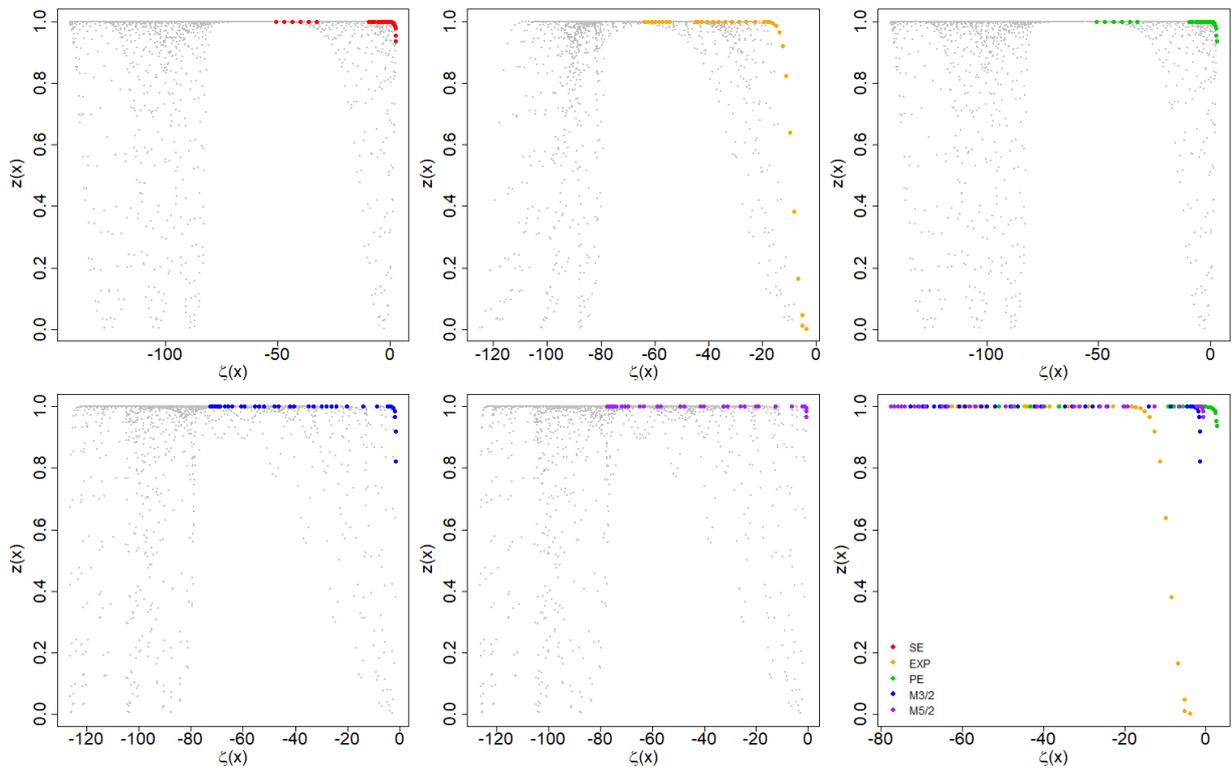

***Figure 5***. *Pareto frontiers obtained by using the distance-based uncertainty quantification (i.e., $u(x) = z(x)$). Five different kernels are used to fit as many GPs, leading to as many Pareto frontiers. Last chart (bottom-right) depicts the five frontiers all together for an easier comparison.*

The only way to analyse how different uncertainty quantification measures can lead to completely different decisions – even if anyway Pareto rational – is to localize, within the search space $\Omega \subset \Re^d$, the locations whose associated objectives lays on the Pareto frontier (namely, the Pareto set). Figure 6 reports just a 2D example, considering ten previous decisions (bold black crosses), five different kernels and three alternative uncertainty quantification measure. The black box function considered is Branin-Hoo.

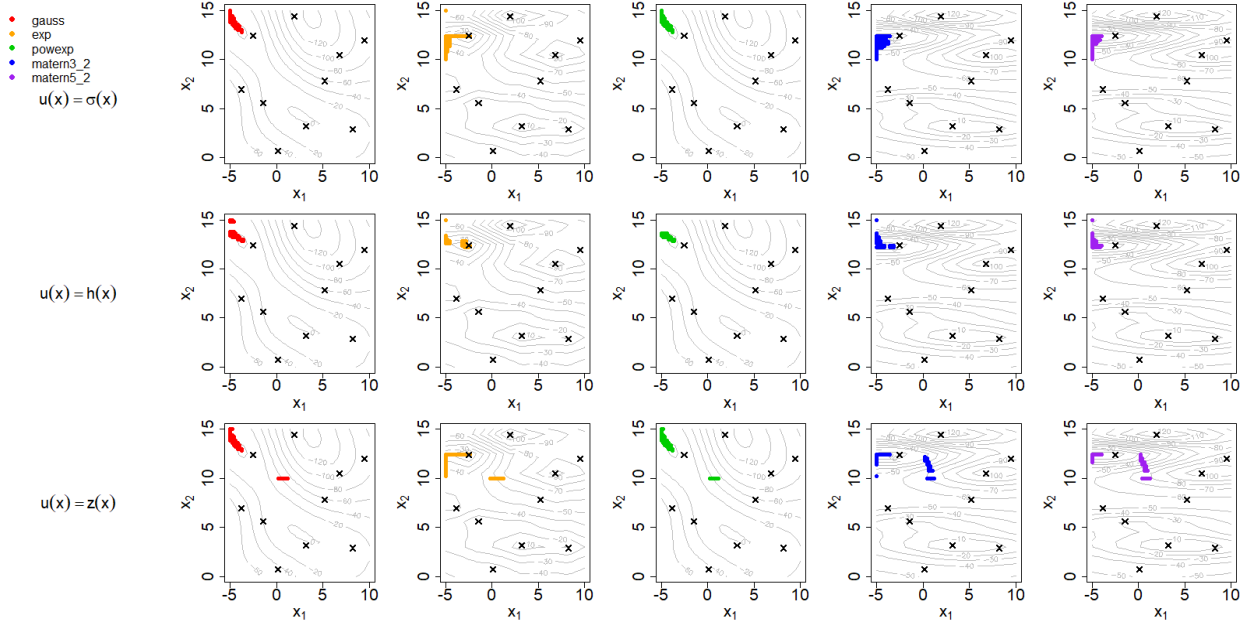

***Figure 6.*** *Next decision depending on: (i) ten previous observations (bold crosses), (ii) uncertainty quantification measures (rows: $\sigma(x)$, $h(x)$ and $z(x)$), and kernels (columns: "gauss" for $k_{SE}$, "exp" for $k_{EXP}$, "powexp" for $k_{PE}$, "matern3_2" for $k_{M3/2}$ and "matern5_2" for $k_{M5/2}$)*

From the figure it is possible to notice that the region of locations associated to Pareto-rational decisions does not change so much depending on kernel, as well as by using $\sigma(x)$ or $h(x)$. The most evident difference arises by using $z(x)$ as uncertainty quantification measure, because it allows to consider as Pareto-rational also decisions in the area around, approximately, the location $(x_1 = 1; x_2 = 11)$. This area is associated to a more explorative behaviour compared to the other – which is also identified by using the other two uncertainty quantification measures – meaning that some explorative choice could be still considered Pareto rational when $u(x) = z(x)$.

This is just an example for explanatory purposes, the hypothesis is investigated and validated in our analysis. Moreover, we have also to consider that humans, (Kahneman, 2011) could take non-Pareto-rational decisions, and it is therefore important to measure how much a decision can be considered *"far from a Pareto-rational one"*. This issue is addressed and formalized in the next section.

### 3.3 Distance from the Pareto rationality

Every next decision, $x^{(n+1)}$, can be analysed according to the distance of its "image" $\left(\zeta(x^{(n+1)}), u(x^{(n+1)})\right)$ from the Pareto frontier, computed as follows:

$$d(\bar{\psi}, \bar{\mathcal{P}}) = \min_{\psi \in \bar{\mathcal{P}}} \left\{ \|\bar{\psi} - \psi\|_2^2 \right\}$$

where $\bar{\psi} = \left(\zeta(x^{(n+1)}), u(x^{(n+1)})\right)$ and $\bar{\mathcal{P}} = \mathcal{P}(\Psi_{1:m}) \cup \{\bar{\psi}\}$.

This distance is computed for every choice among the five kernels and the three uncertainty quantification measures previously presented. The hypothesis is that humans are mostly Pareto rational, and they should therefore make decisions laying on – or close to – the Pareto frontier.

We analyse the distances from all the 15 possible Pareto frontiers (5 kernels × 3 uncertainty quantification measures) and how they change along the optimization process. Figure 7 shows an example taken from our experimental results and anticipated here just for explanatory purposes. The 10 charts refer to as many black box optimization problems solved by a single human subject. At each iteration, and for each uncertainty quantification measure, the minimum distance between the associated Pareto frontier and the human decision is reported, irrespectively to the kernel. It is important to remark that distances cannot be compared in absolute terms, because the three different uncertainty measures can vary in very different ranges. However, it is possible to observe that distances result correlated in some cases and uncorrelated in others.

Finally, from the charts it is possible to notice that: *(a)* in some cases Pareto rationality is independent on the uncertainty quantification, such as for the problems: *bukin6*, *goldpr*, *rastr*, *stybtang*, but not in general; *(b)* a higher number of decisions are considered Pareto rational if $u(x) = z(x)$; *(c)* in some cases it is possible to observe a shift from Pareto-rationality to *not*-Pareto rationality (e.g., this is evident in for *beale, goldpr* and *rastr*).

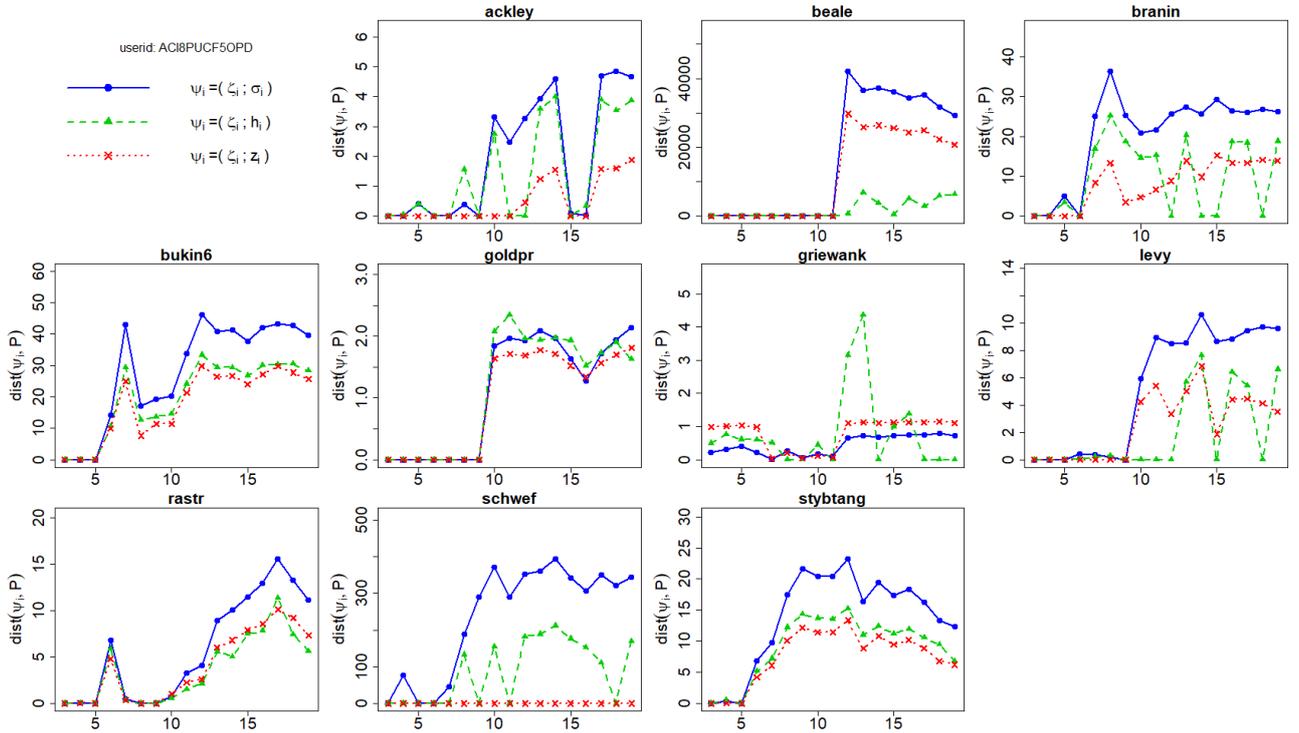

***Figure 7.*** *Distance, at each iteration, of the next decision from three different Pareto frontiers, one for each uncertainty quantification measure $\sigma(x)$, $h(x)$ and $z(x)$. All the fifteen charts are related to as many black box optimization tasks performed by a single human subject.*

## 4. Experimental setup

### 4.1 Data collection

To collect data about humans' strategies we have used a gaming application based on the implementation used in (Candelieri et al., 2020). Figure 8 shows the web-based Graphical User Interface (GUI) of our game, with a game play example. The game field, with previous decisions and observations, as well as the score and remaining "shots", are reported.

The game can be arranged according to different goals/conditions.
- **Game mode #1**: searching for the location having the highest score;
- **Game mode #2**: searching for the location having the highest score, but given the additional information about its value;
- **Game mode #3**: maximizing the cumulative score (sum of the scores of all the choices).

This paper focuses only on the first game mode (i.e., Game mode #1), with the aim to formalize and validate an analytical framework which could be successively adopted to investigate how the different goals of the other two game modes can imply different the humans' strategies.

Fourteen volunteers have been enrolled (among families and friends, which had no competences in computer science and/or optimization), asking for solving ten different tasks each (only for Game mode #1). Each task refers to a global optimization test function, which subjects "learn and optimize" by clicking at a location and observing the associated score (aka reward). For each task, every player has a maximum number of 20 clicks (decisions) available. The 10 global optimization test functions adopted are depicted in Appendix (A1). Since these functions are related to minimization problems, the score returned to the player is $-f(x)$, translating them into maximization tasks.

Finally, the game has been developed in R, specifically R-shiny for the web-based GUI. All the analytical components, described in the following, have been also developed in R as backend of the application.

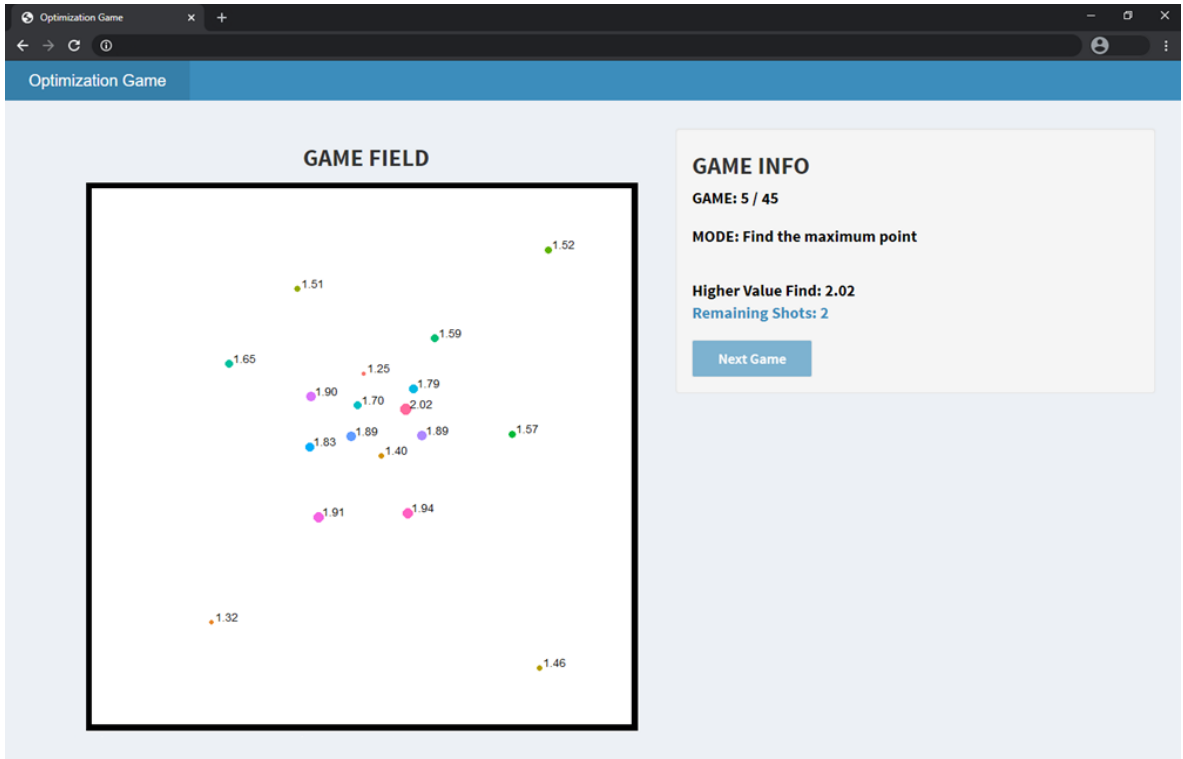

*Figure 8. Web-based Graphical User Interface (GUI) of our game: a game play example*

**4.2 Data analysis**

This study analyses every single decision performed by every volunteer, and how it can be explained in terms of uncertainty quantification and Pareto rationality. The analysis is organized in three consecutive steps:

- *Step 1:* computing the number of Pareto rational decisions, depending on the uncertainty quantification measure, and comparing them. A decision is considered Pareto rational if the distance from the Pareto frontier is less than $10^{-4}$. This analysis step is summarized as follows:
    1. For each player and each test problem do:
    2. Initialize $n = 3$ (i.e., he first three decisions of each user and for each test problem cannot be analysed, because fitting a GP over a 2-dimensional search space requires at least three observations).
    3. Condition a GP for each one the kernels in $\mathcal{K}$ to the previous $n$ decisions and observations performed by that player for that test problem.
    4. For each $u(x) \in \mathcal{U}$ do:

5. Use each one of the conditioned GPs to approximate the associated Pareto frontier by sampling a grid of $m = 30 \times 30 = 900$ locations in $\Omega$ and then computing the associated locations in the $\zeta$-$u$ space.
6. Map $x^{(n+1)}$ into five associated locations $\psi = \left(\zeta(x^{(n+1)}), u(x^{(n+1)})\right)$ – one for each kernel – and compute its minimum distance from the five Pareto frontiers and store it.
7. $n \leftarrow n + 1$ and go to Step 3
8. End For each $u(x) \in \mathcal{U}$
9. End For each player and each test problem

All the results are stored into a data table with columns: *user_id*, *problem_id*, *n+1*, *uncertainty_measure*, *min_dist_from_Pareto_frontier*. Finally, numbers of Pareto rational decisions are separately computed for each uncertainty quantification measure and aggregated by *(a)* players and *(b)* test problems.

- *Step 2:* computing the length of consecutive Pareto-rational decisions, depending on the uncertainty quantification measure, and comparing them. This analysis step uses the same data table previously computed but, instead of the number, the length of consecutive Pareto rational decisions is computed, separately for the three uncertainty measures, and aggregated by *(a)* players and *(b)* test problems.

- *Step 3:* depending on results from the two previous steps, the uncertainty quantification measure which allows to more frequently classify the humans' choices as Pareto rational is selected. Then, the relationship between the fact that the decision is Pareto rational and the *reward* collected so far by the user is investigated, with the aim to identify a possible motivation for *over-explorative* decisions (i.e., not-Pareto rational decisions). In our analysis *reward* is represented by the score immediately observed by the player implied by his/her own decision. *Cumulative reward* is therefore the sum of scores collected up to a given decision. Finally, the *Average Cumulated Reward* (ACR), up to a given decision, is computed as the arithmetic mean of the cumulated reward up to that decision:

$$ACR^{(n+1)} = \frac{1}{n}\sum_{i=1}^{n} y^{(i)}$$

we use $ACR^{(n+1)}$ to denote the average reward collected up to $n$ just to be coherent with indexing, since this value is analysed in relation with the Pareto distance of decision $x^{(n+1)}$. The idea is that ACR could quantify the amount of "gratification" (high values of ACR) or "stress" (low values of ACR) experienced by the player in solving the test problem. The hypothesis is that lower values of ACR could be associated to not-Pareto rational decisions, induced by a sense of stress for the incapability to (further) improve the score.

## 5. Experimental results and their analysis

### 5.1 Results about analysis step 1

The main result from analysis step 1 is that using $z(x)$ as uncertainty quantification measure increases the number of Pareto rational decisions for some problems, over all the players. Figure 9 shows the number of players with respect to the percentage of Pareto rational decisions. A stacked histogram is provided for each test problem, comparing the distributions obtained for each one of the three uncertainty quantification measures. The increase in terms of number of Pareto rational decisions, by using $u(x) = z(x)$, is more evident for the test problems *schwef* and *ackley*.

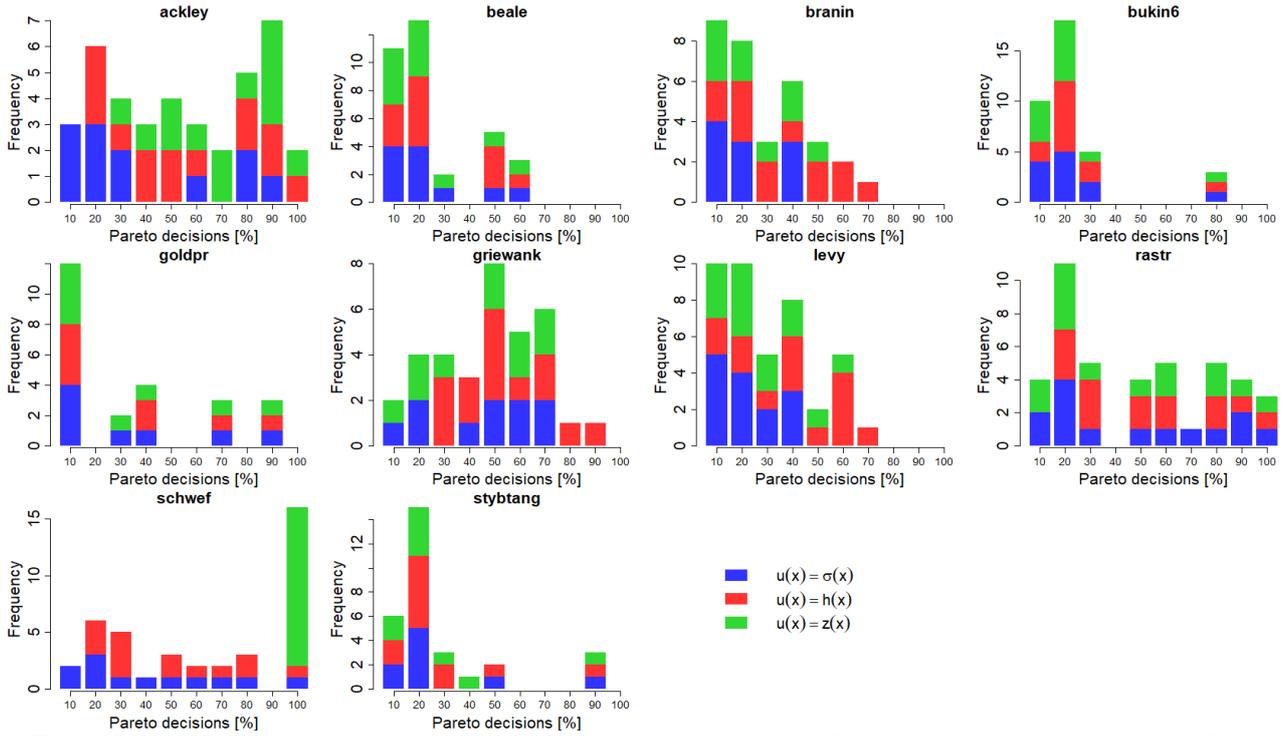

*Figure 9. Number of players with respect to percentage of decisions classified as Pareto rational, separately for the three uncertainty quantification measures. One chart for each test problem.*

Moreover, the higher number of Pareto rational decisions, obtained by using $u(x) = z(x)$, is spread over all the players. Indeed, Figure 10 shows the distributions of the number of test problems with respect to the percentage of Pareto rational decisions. A stacked histogram is provided for each player, comparing the distributions obtained considering each one of the three uncertainty quantification measures. For almost all the subjects a higher number of test problems is solved by using a high percentage of Pareto rational decisions when $u(x) = z(x)$.

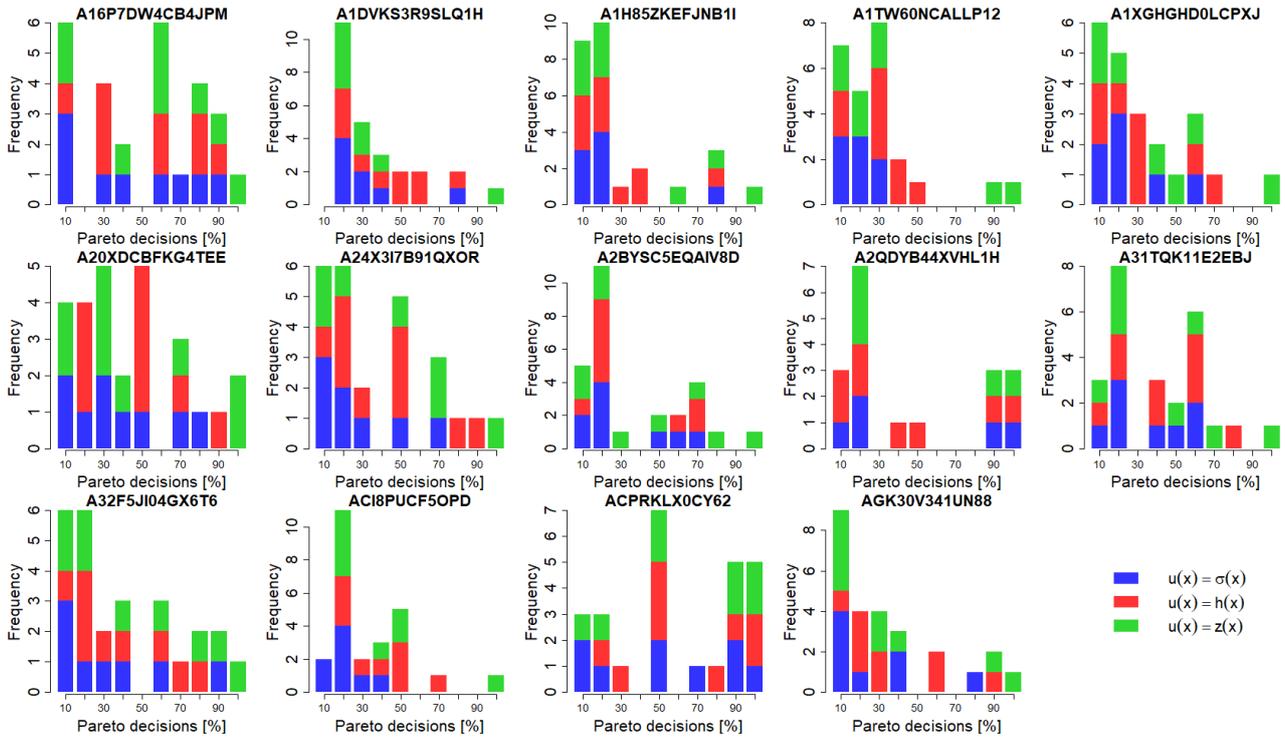

*Figure 10. Number of test problems with respect to percentage of decisions classified as Pareto rational, separately for the three uncertainty quantification measures. One chart for each player.*

## 5.2 Results about analysis step 2

Results of analysis step 2 confirm those from the previous step. Choosing $u(x) = z(x)$ leads to longer sequences of consecutive Pareto rational decisions, according to both the number of players for each test function (Figure 11) and the number of test functions for each player (Figure 12).

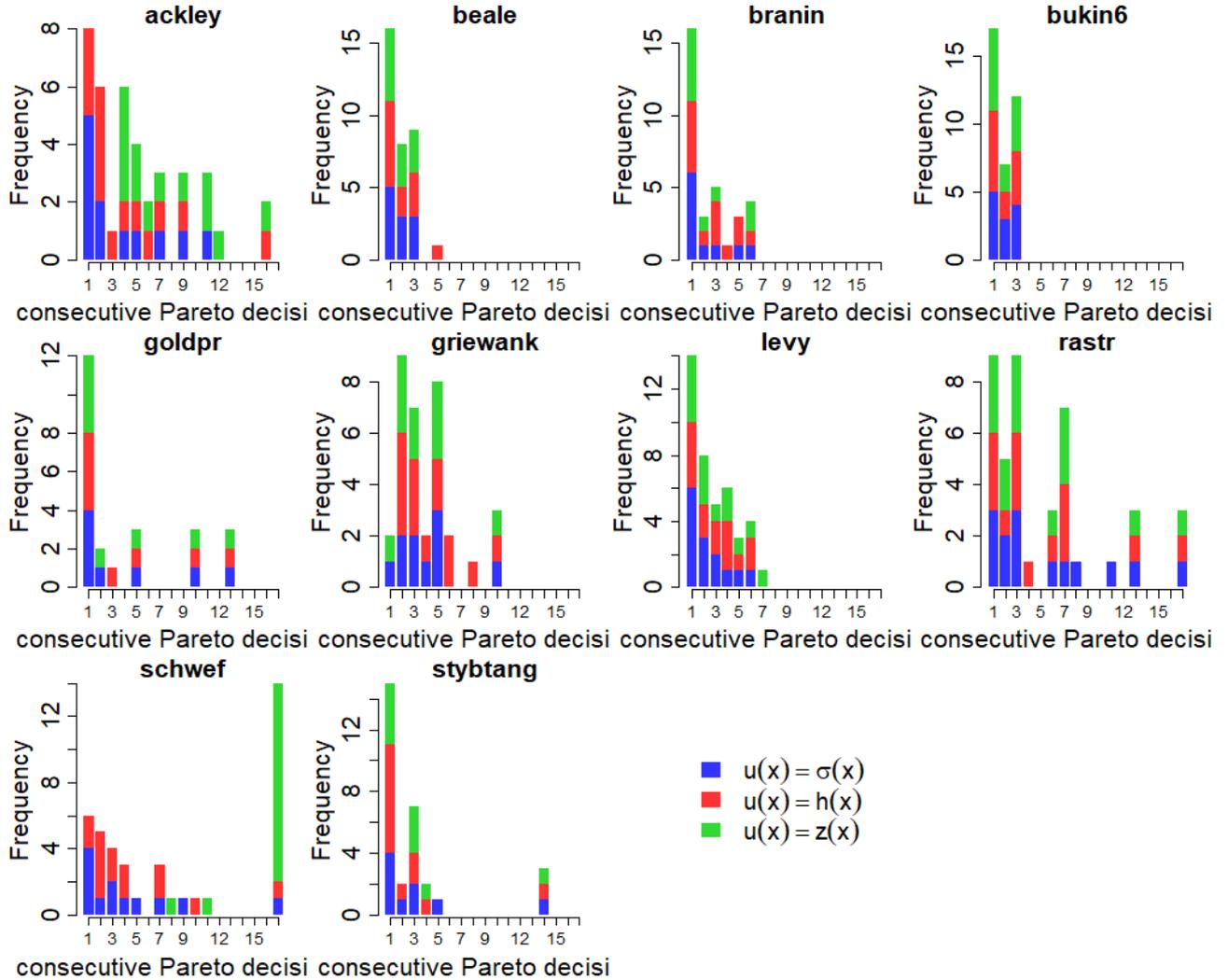

*Figure 11. Number of players with respect to length of consecutive Pareto rational decisions, separately for the three uncertainty quantification measures. One chart for each test function.*

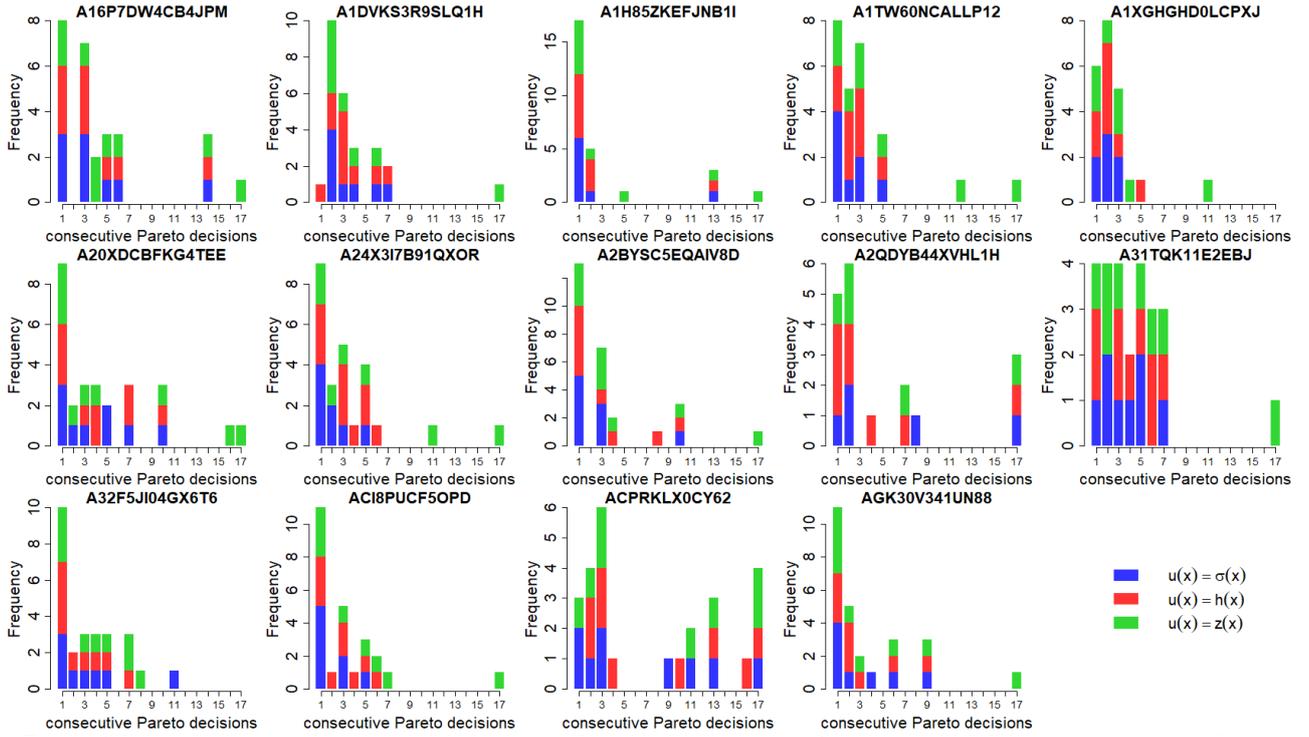

***Figure 12**. Number of test functions with respect to length of consecutive Pareto rational decisions, separately for the three uncertainty quantification measures. One chart for each player.*

## 5.3 Results about analysis step 3

According to the results from the two previous steps, we can conclude that $z(x)$ is, among the uncertainty quantification measures considered, the one inducing the Pareto model with the highest representation power, that is the one maximizing the number of Pareto optimal decisions. Indeed, if we assume that every human's decision is Pareto rational, then $z(x)$ is the only option – at least among those considered – allowing us to get close to this rationality model. Therefore, we have selected $u(x) = z(x)$ to perform the analysis step 3.

As the main result of this step, the value of the ACR can help to determine if the next decision $x^{(n+1)}$ will be Pareto rational or not. More specifically, ACR resulted, on average, higher in the case of a Pareto rational decision on 8 out of the 10 test problems (in 4 cases, this difference is statistically significant, *p*-value<0.05, U Mann-Whitney test). Only in one case (i.e., *stybtang*) the ACR is significantly higher for not-Pareto decisions (*p*-value<0.001, U Mann-Whitney test). Results are reported in Table 1 and, for a more immediate comparison, also as boxplots in Figure 13.

***Table 1** Results: comparing ACR between Parto and not-Pareto rational decisions. Results are per test function, over all players and decisions.*

| test function | ACR Pareto mean (sd) | ACR not-Pareto mean (sd) | U Mann-Whitney test *p*-value |
|---|---|---|---|
| ackley | **-178.916** (95.209) | -188.259 (96.440) | 0.409 |
| beale | -54603.570 (111582.400) | **-53695.820** (111111.300) | 0.170 |
| branin | **-209.413** (224.193) | -380.112 (268.029) | *<0.001\** |
| bukin6 | **-482.495** (203.231) | -995.122 (473.103) | *<0.001\** |
| goldpr | **-20.601** (21.869) | -24.551 (14.557) | *0.030\** |
| griewank | **-7.791** (4.122) | -8.485 (5.610) | 0.9015 |
| levy | **-92.548** (69.508) | -114.750 (88.185) | 0.1363 |
| rastr | **-309.380** (145.268) | -421.085 (178.801) | *<0.001\** |
| schwef | **-8859.593** (3905.512) | -9426.500 (2749.823) | 0.816 |
| stybtang | 157.231 (100.050) | **319.434** (227.555) | *<0.001\** |

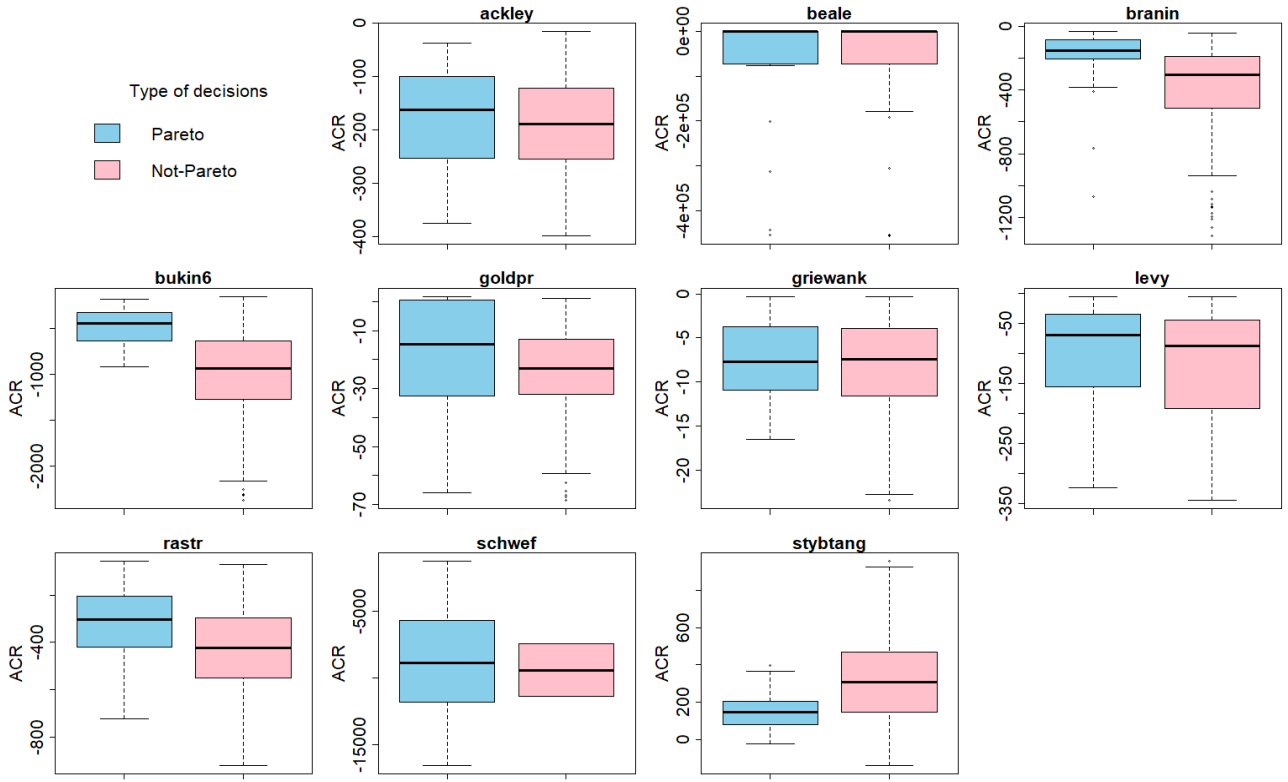

*Figure 13. Comparing ACR between Pareto and not-Pareto rational decisions: a boxplot for each test problem, data aggregated over the decisions of all the players.*

## 6. Conclusions and perspectives

The main result of this paper is a methodological framework to collect and analyse data related to humans' decision-making strategies under uncertainty, specifically about how they balance gathering new information (exploration) and reward seeking (exploitation). To better model this balance, we have used a bi-objective setting and assumed that humans' choices might be more frequently located on the Pareto frontier (Pareto rational choices). Since one of the two objectives is uncertainty, we have analysed three uncertainty quantification measures to investigate which one would offer the best fit with the Pareto rationality model (i.e., the one maximizing the number of choices laying on the associated Pareto frontier).

Thus, while most of previous research studies has investigated how people assess the information value of possible queries, we rather addressed the issue of the perception of probabilistic uncertainty itself. This problem is still an open question in Machine Learning and cognitive sciences and neither our results nor those prevailing in the rich literature about this issue provide unequivocal evidence about the underlying algorithms used by humans.

Humans do not always make "rationale" choices (i.e., Pareto optimal decisions in the space of expected improvement and uncertainty) and in some cases, they "exasperate" exploration. The computational results and their analysis allow to formulate at least a tentative answer to why or rather in which conditions we observe deviations from "rationality" and switches towards "exasperated" exploration depending on the dynamics of the optimization process as represented by Average Cumulative Reward.

Next steps should be a probabilistic characterization of the sequence of decisions and a close analysis of the dynamics of how people change their behaviour. A big question, which we have only partially addressed in this paper, is whether this analysis sits well with the Paretian expected utility theory or should rather be developed along entirely different lines of inquiry as for instance those proposed in (Peters, 2019) bringing about new uncertainty quantification measures and a new family of Bayesian Optimization acquisition functions.

**Acknowledgements**
We greatly acknowledge the DEMS Data Science Lab, Department of Economics Management and Statistics (DEMS), for supporting this work by providing computational resources.

**Conflicts of interest/Competing interests (include appropriate disclosures)**
Authors declare that they do not have any conflicts of interests or competing interests.

**Ethics approval (include appropriate approvals or waivers)**
Informed consent was given in accordance with the Helsinki declaration.

**Availability of data and material (data transparency):**
Both data and code for reproducing analysis and results of this paper are available at the following link: https://github.com/acandelieri/humans_strategies_analysis.

# Appendix A

## A1. The ten test problems

The ten global optimization test functions used in this study, including their analytical formulations, search spaces and information about optimums and optimizers, can be found at the following link:
https://www.sfu.ca/~ssurjano/optimization.html

Since they are minimization test functions, we have returned $-f(x)$ as score in order to translate them into the maximization problems depicted in Figure 14.

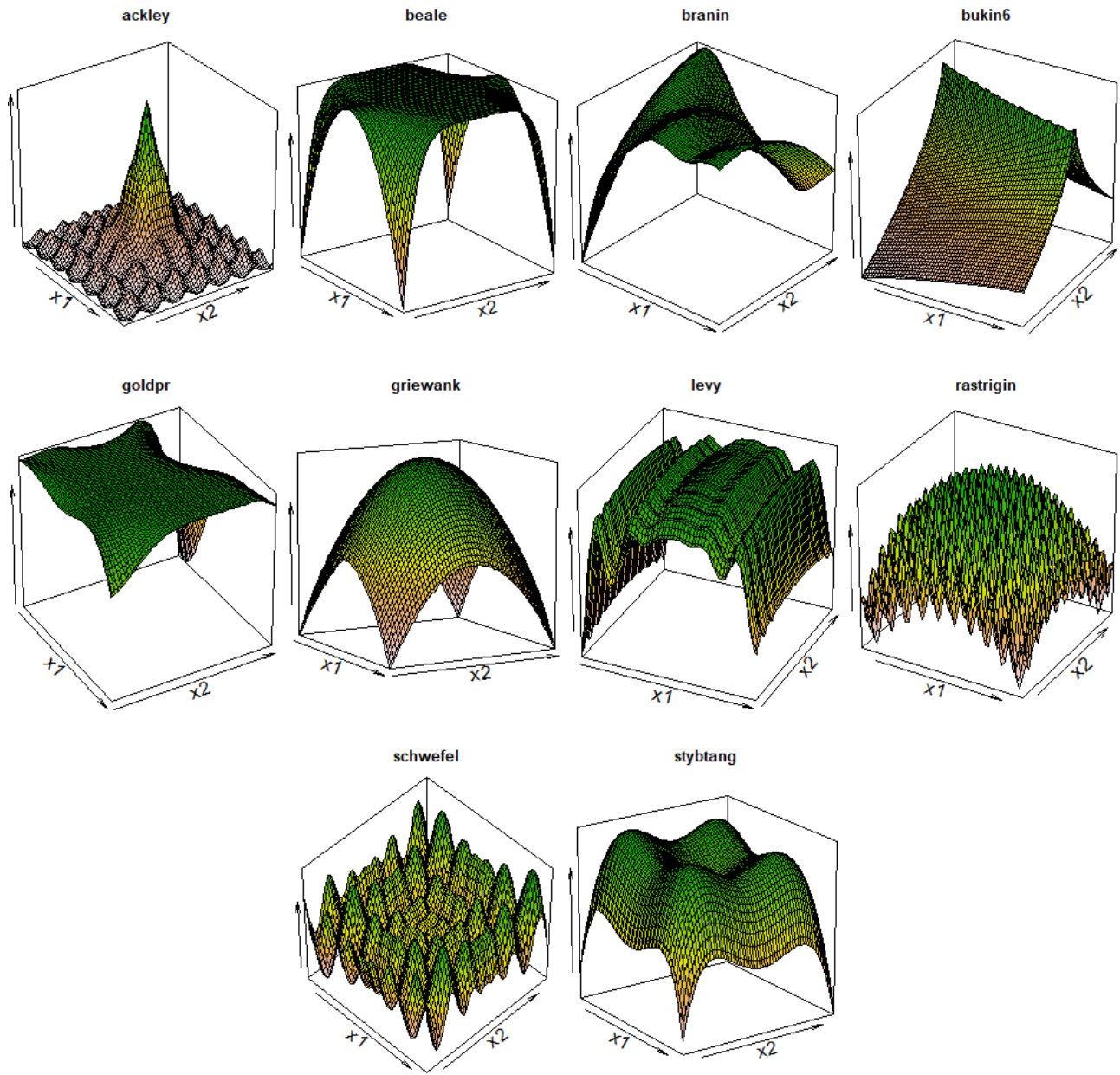

*Figure 14. The 10 test problems considered in this study.*

## A2. Distances from Pareto frontiers for each player, by test function

The following 10 figures – one for each test function – report the distances of each decision from the Pareto frontiers and for each player.

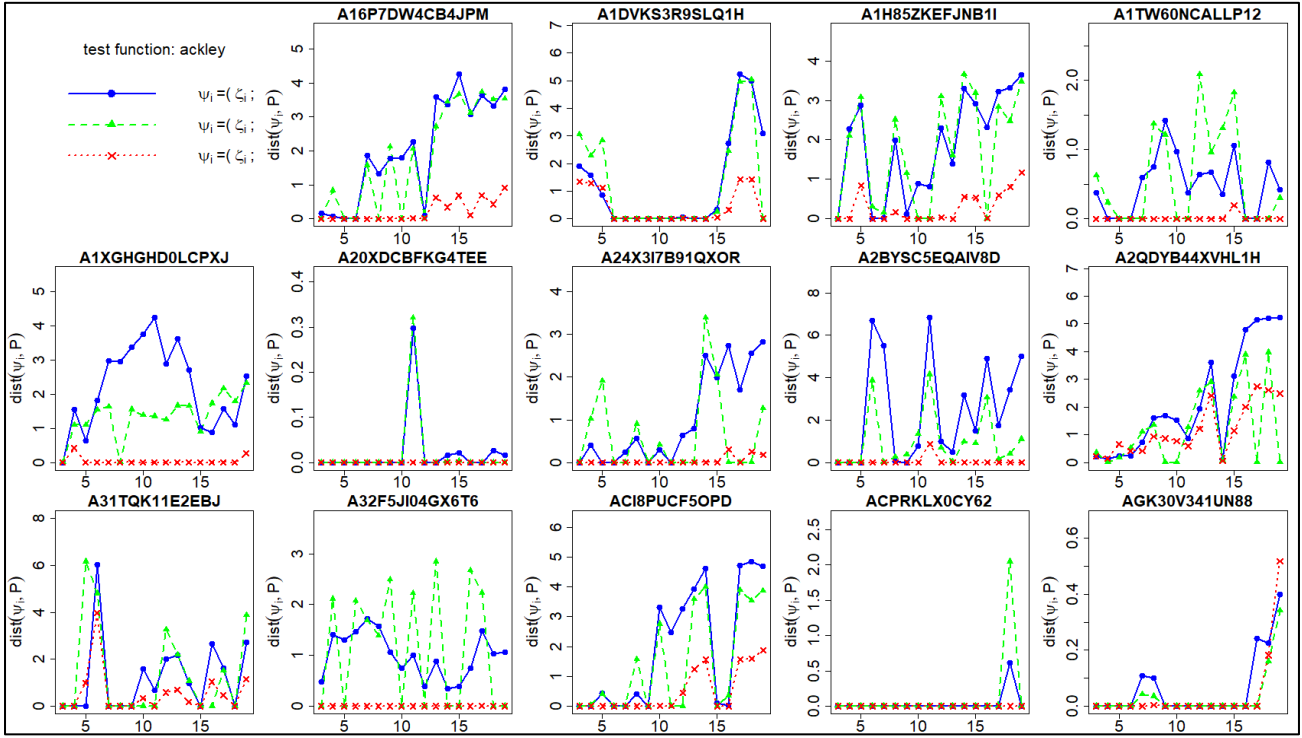

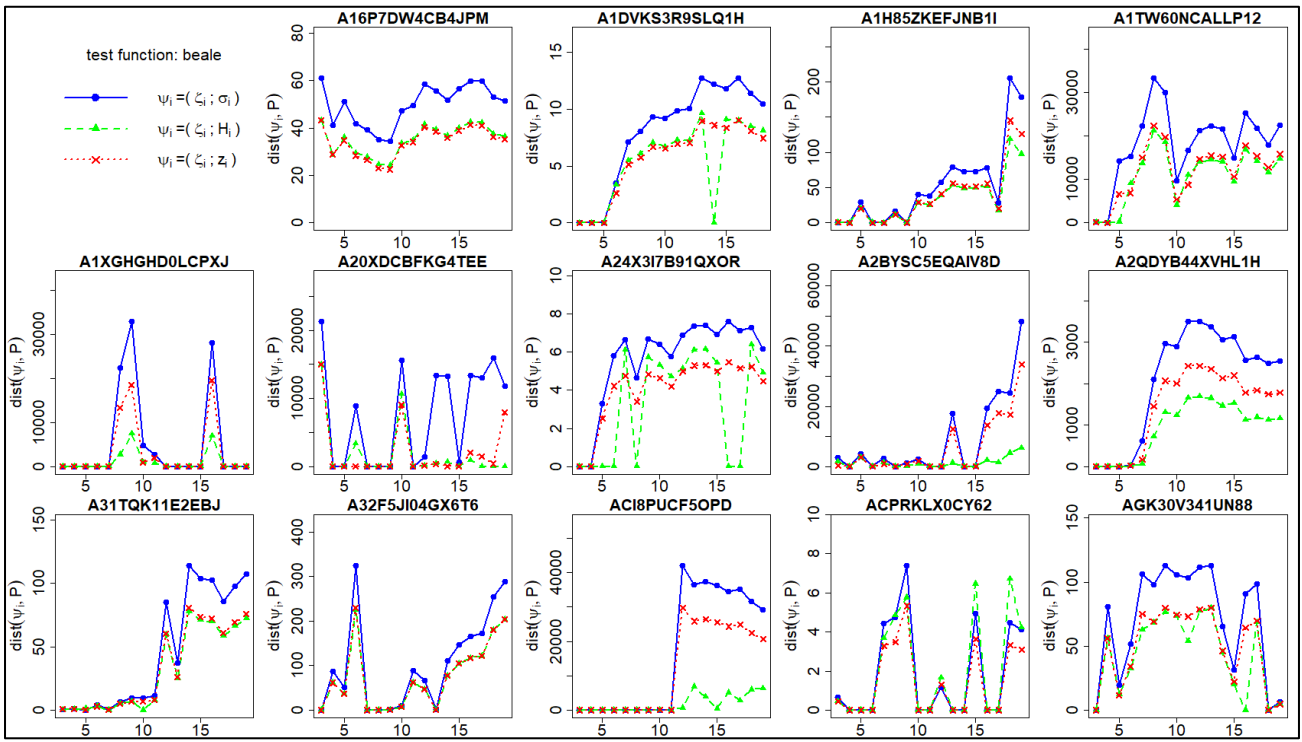

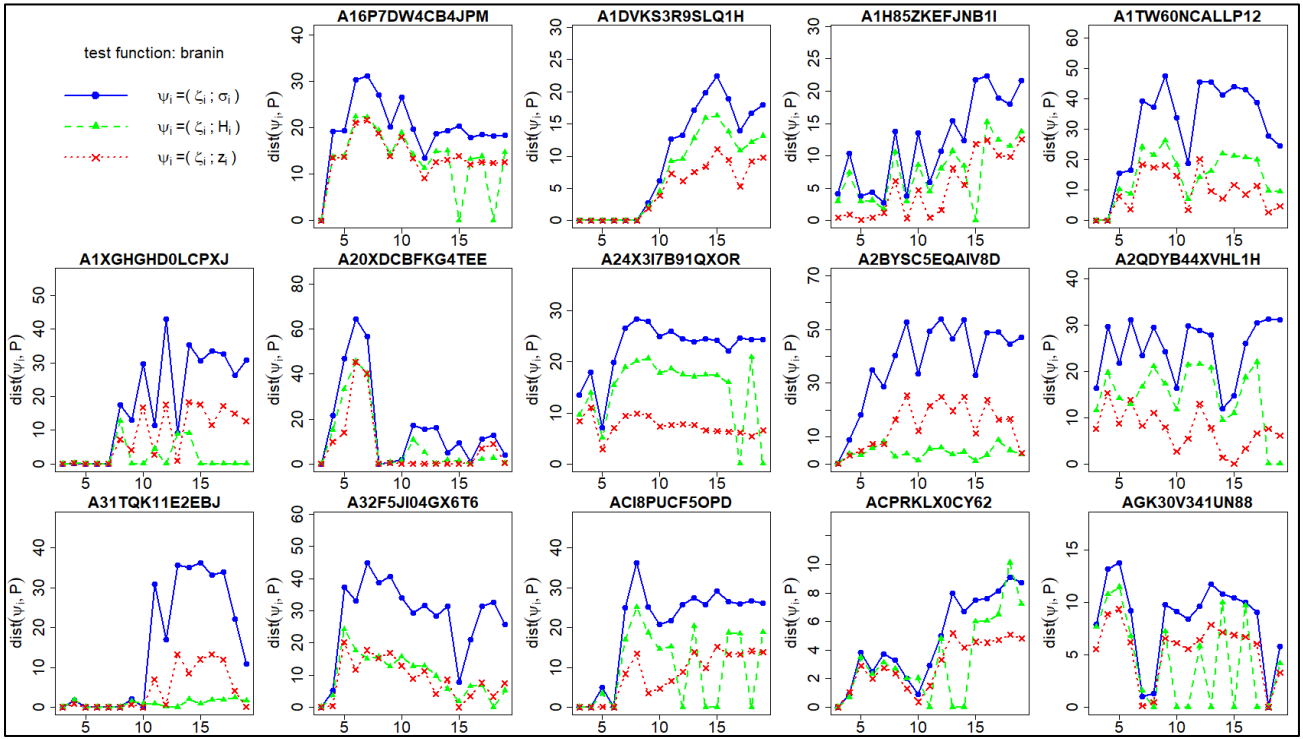

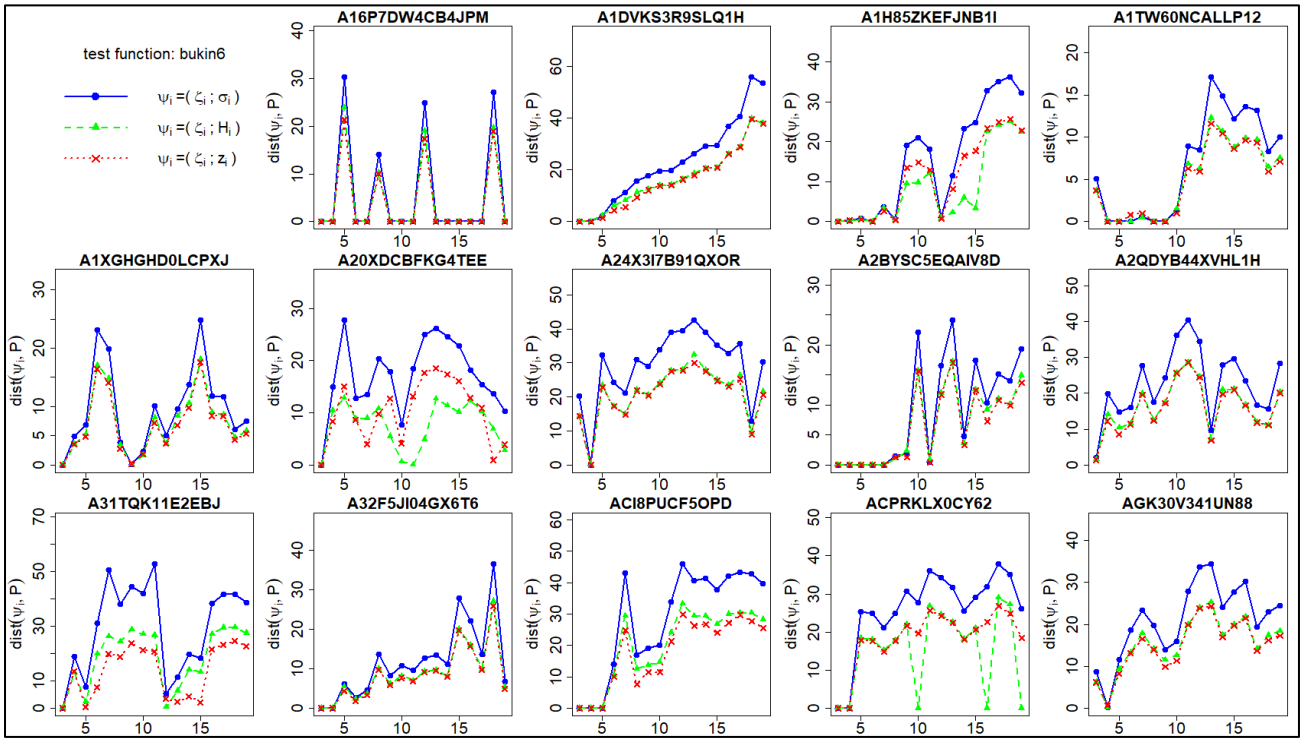

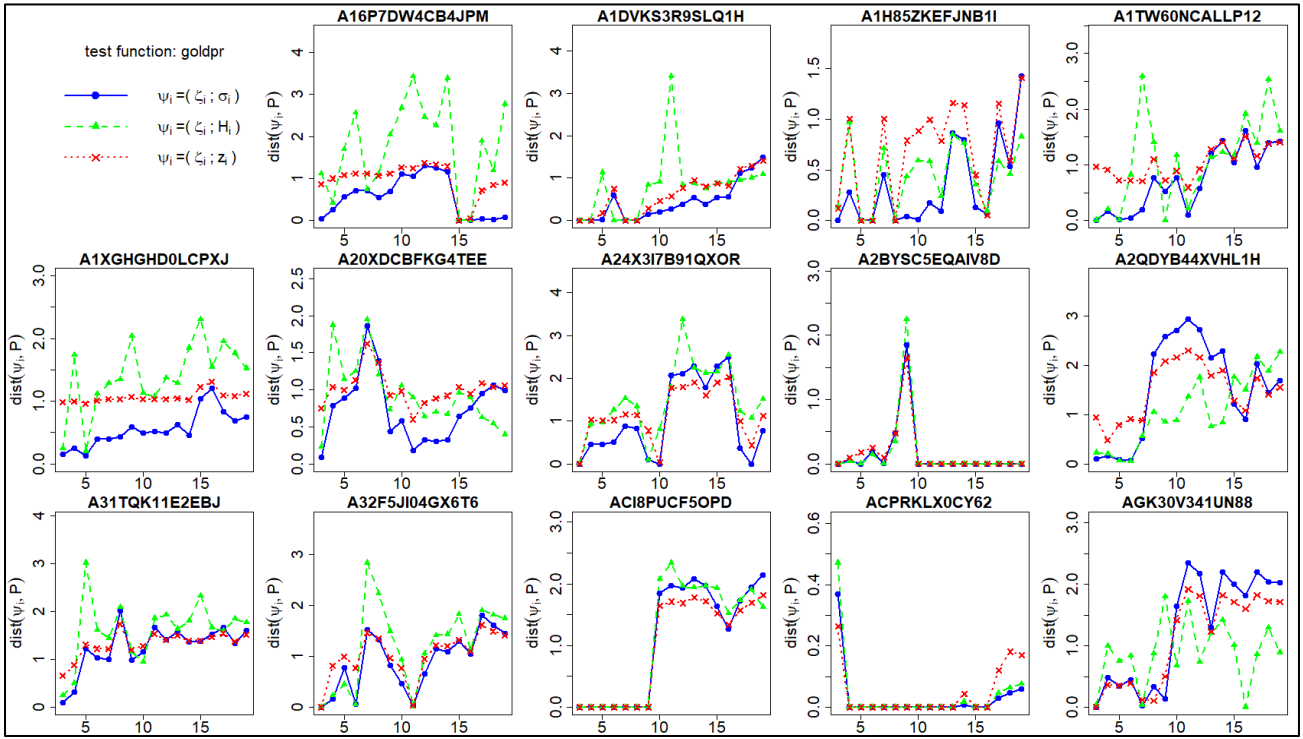
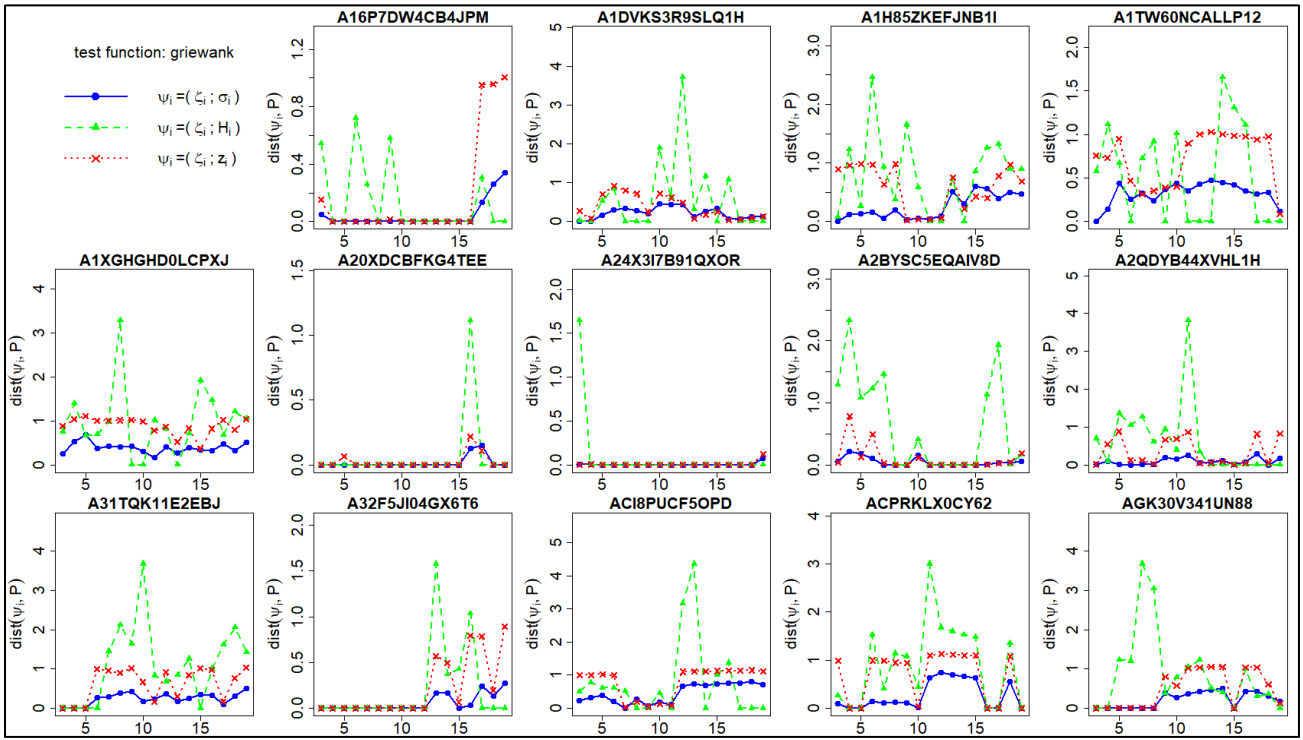

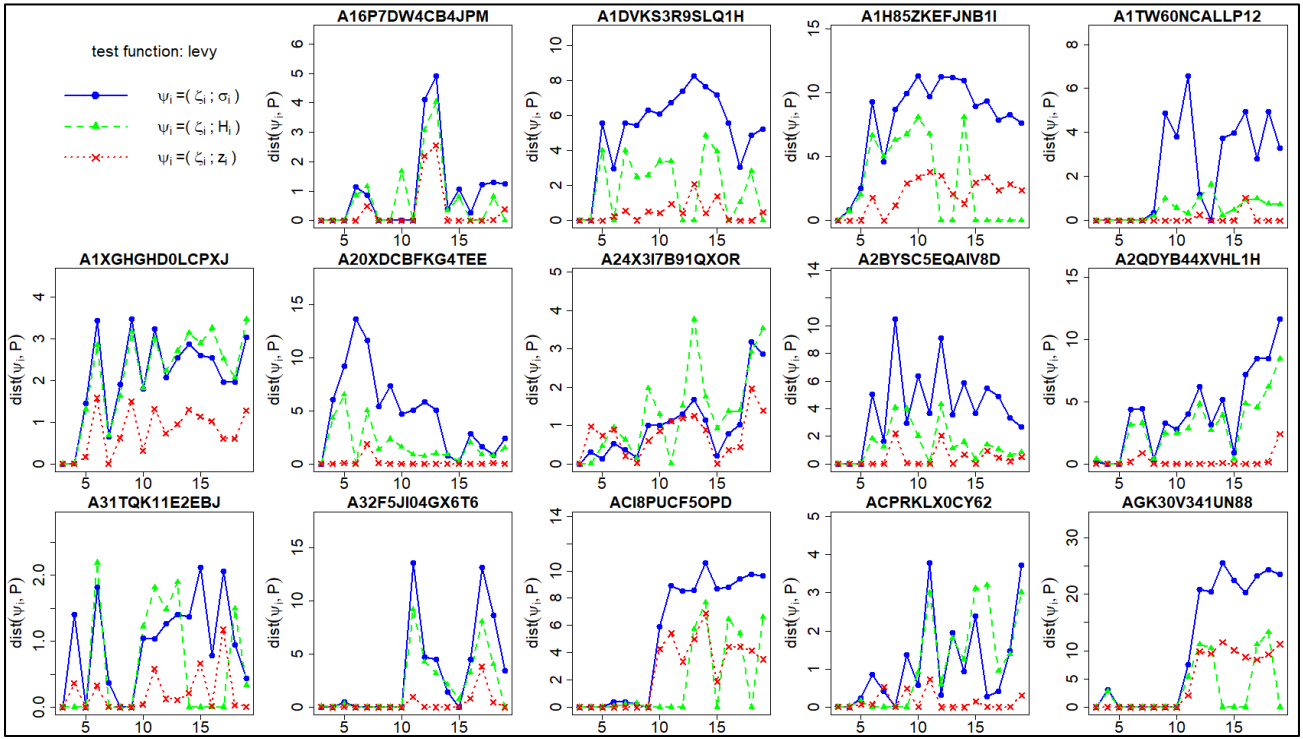

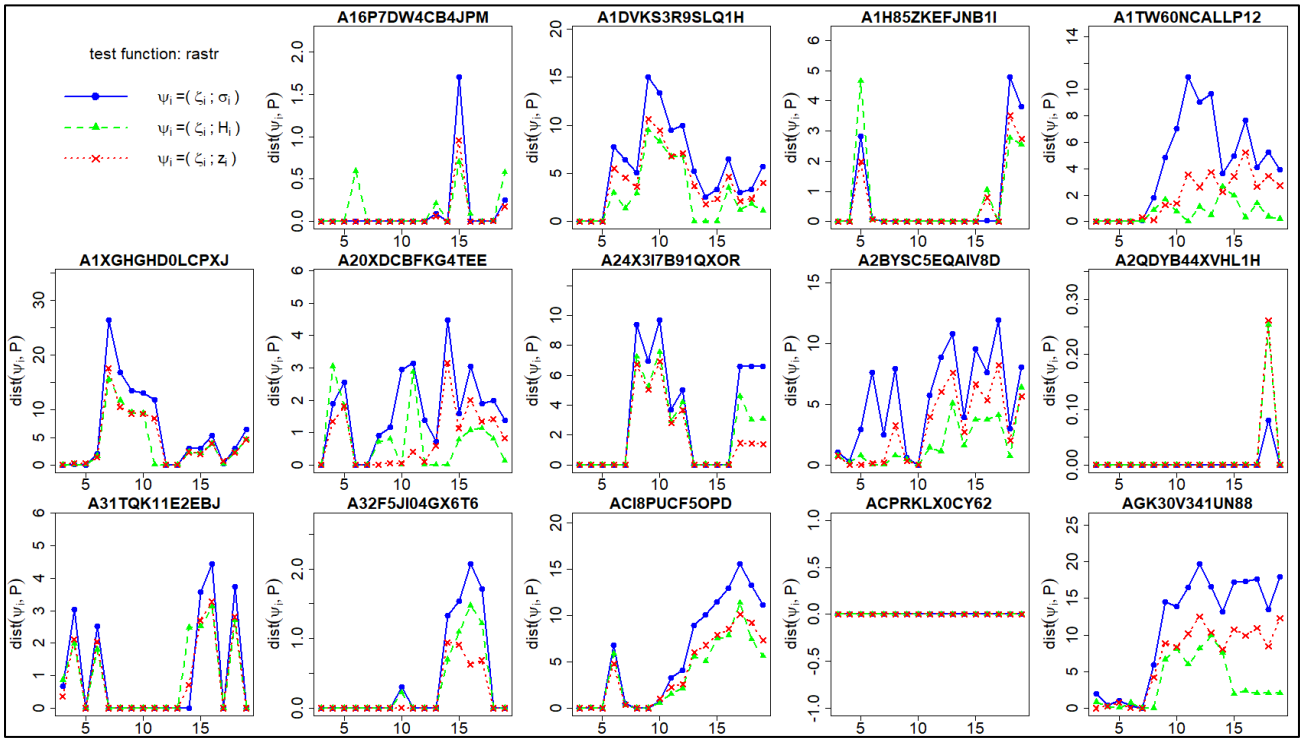

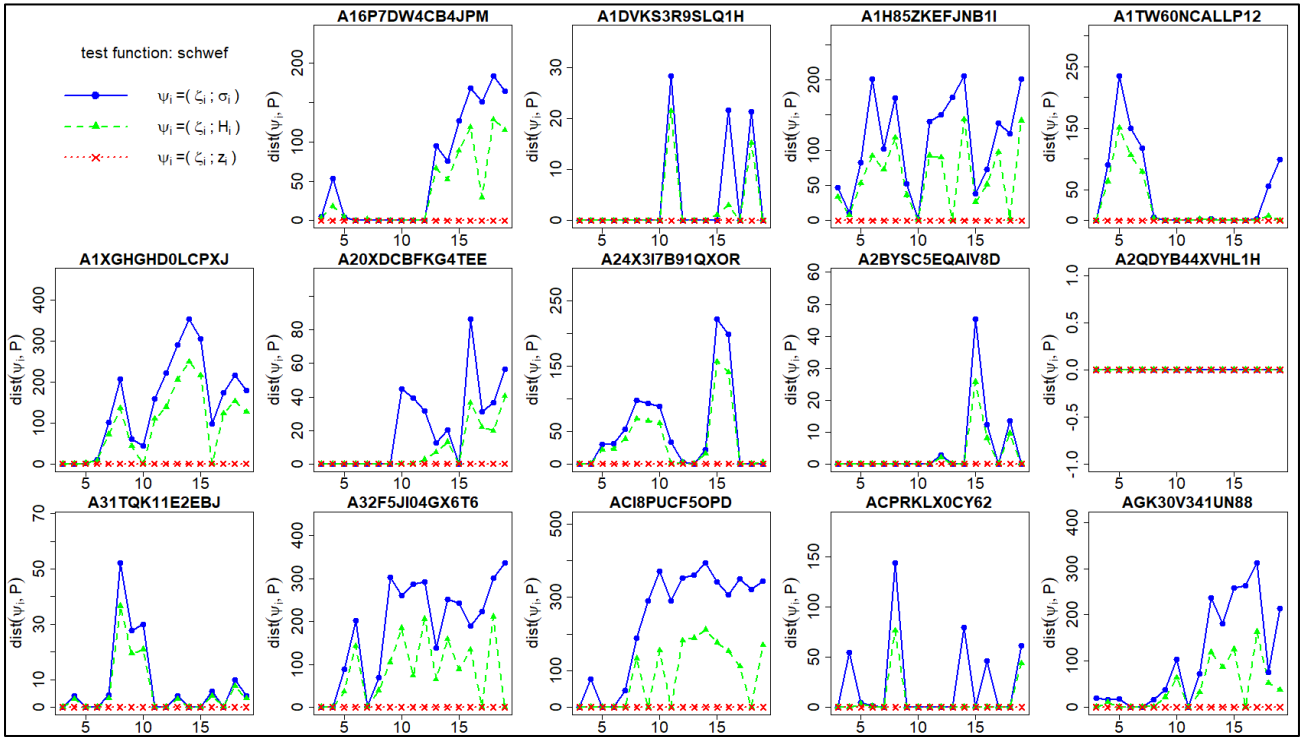
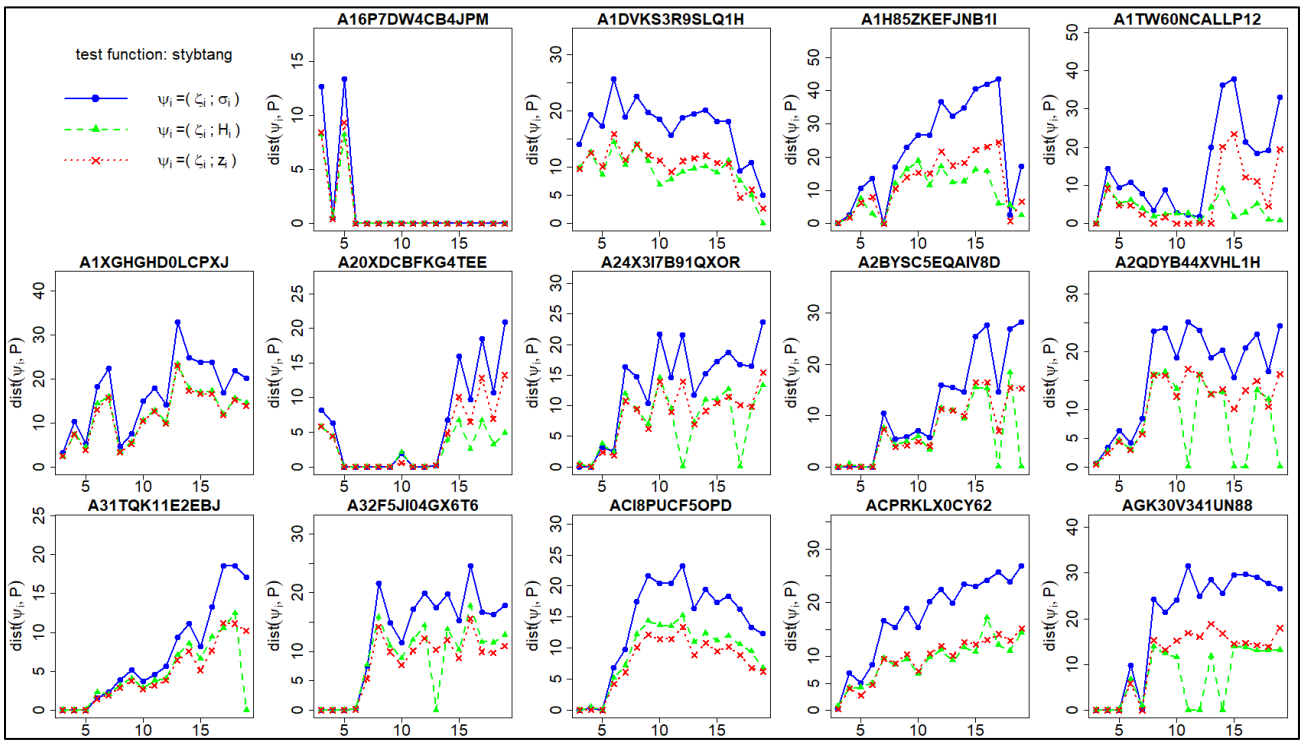

## A3. Distances from Pareto frontiers for each test functions, by player

The following 14 figures – one for each player – report the distances of each decision from the Pareto frontiers and with respect to each test function.

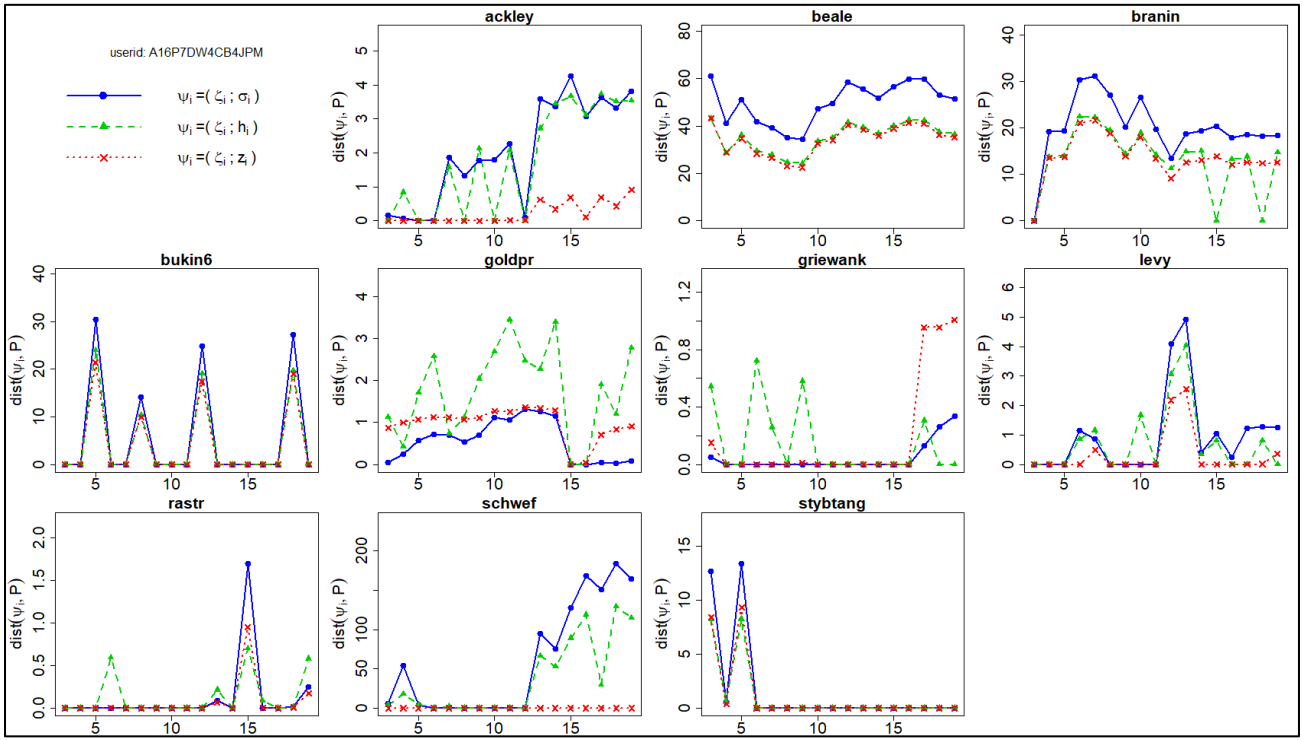

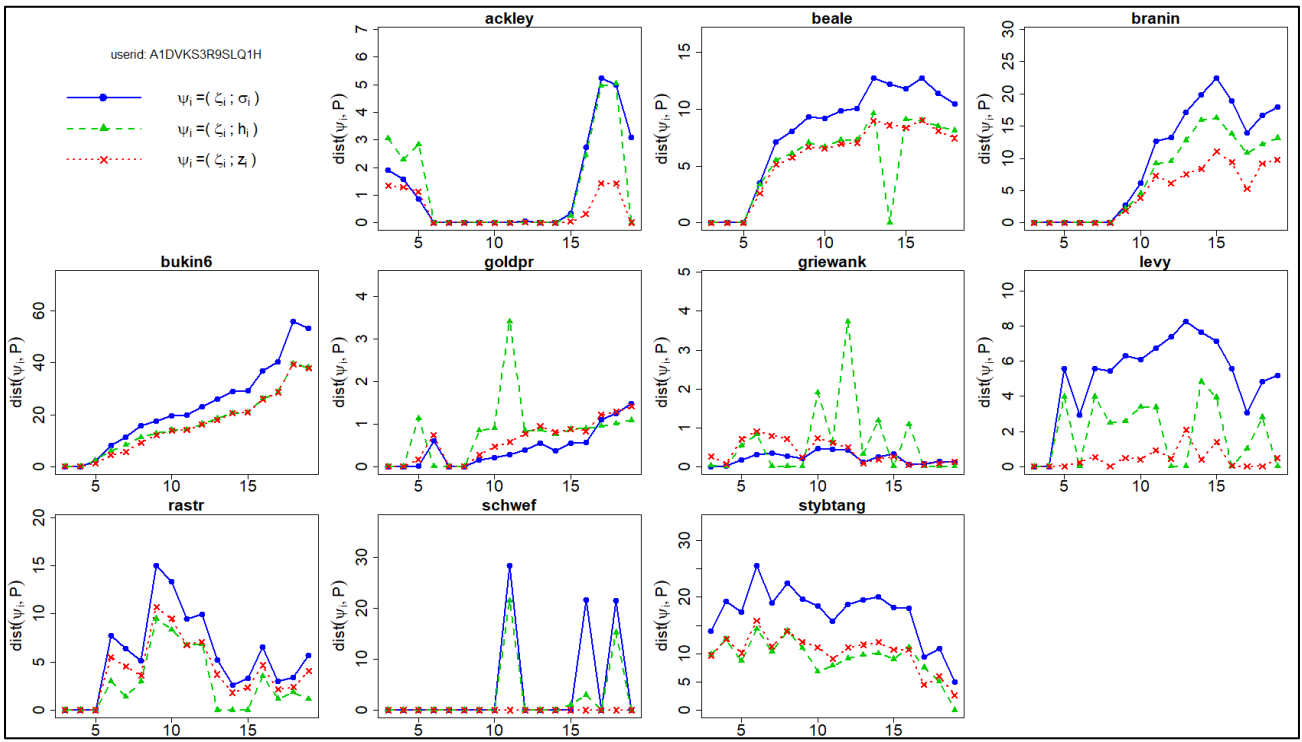

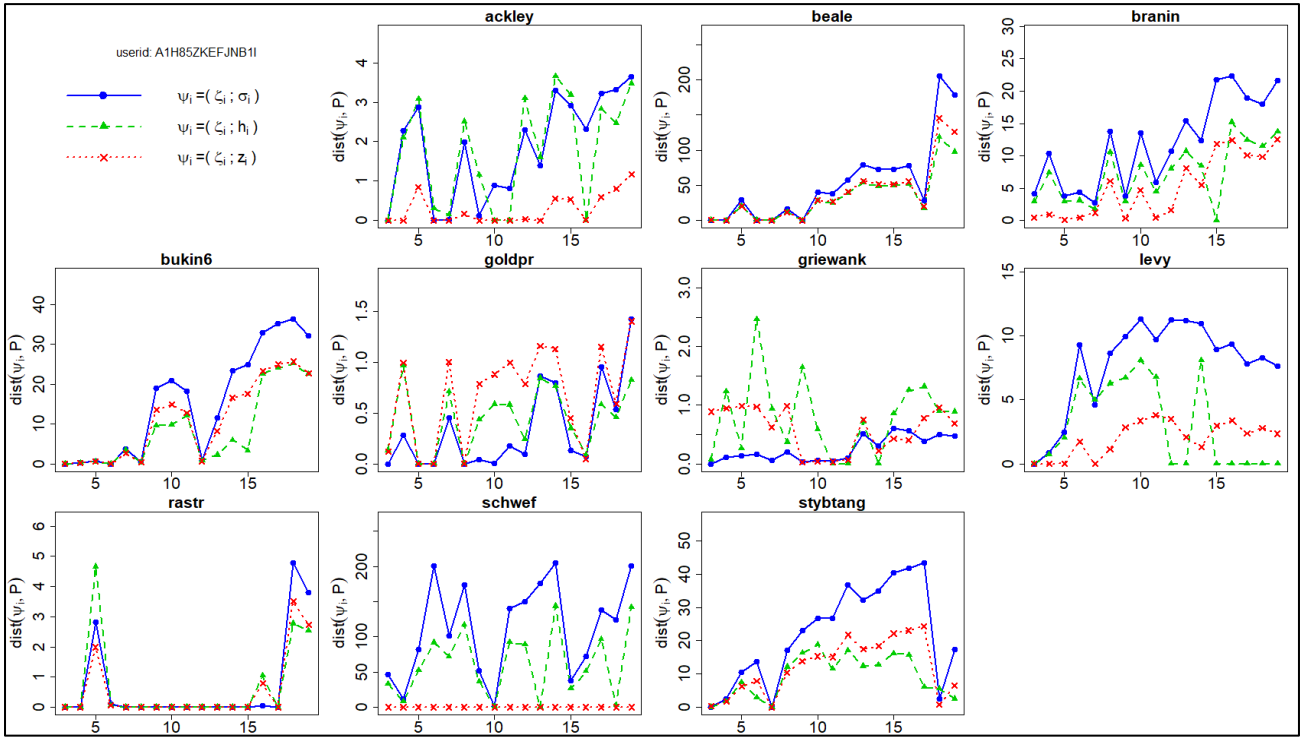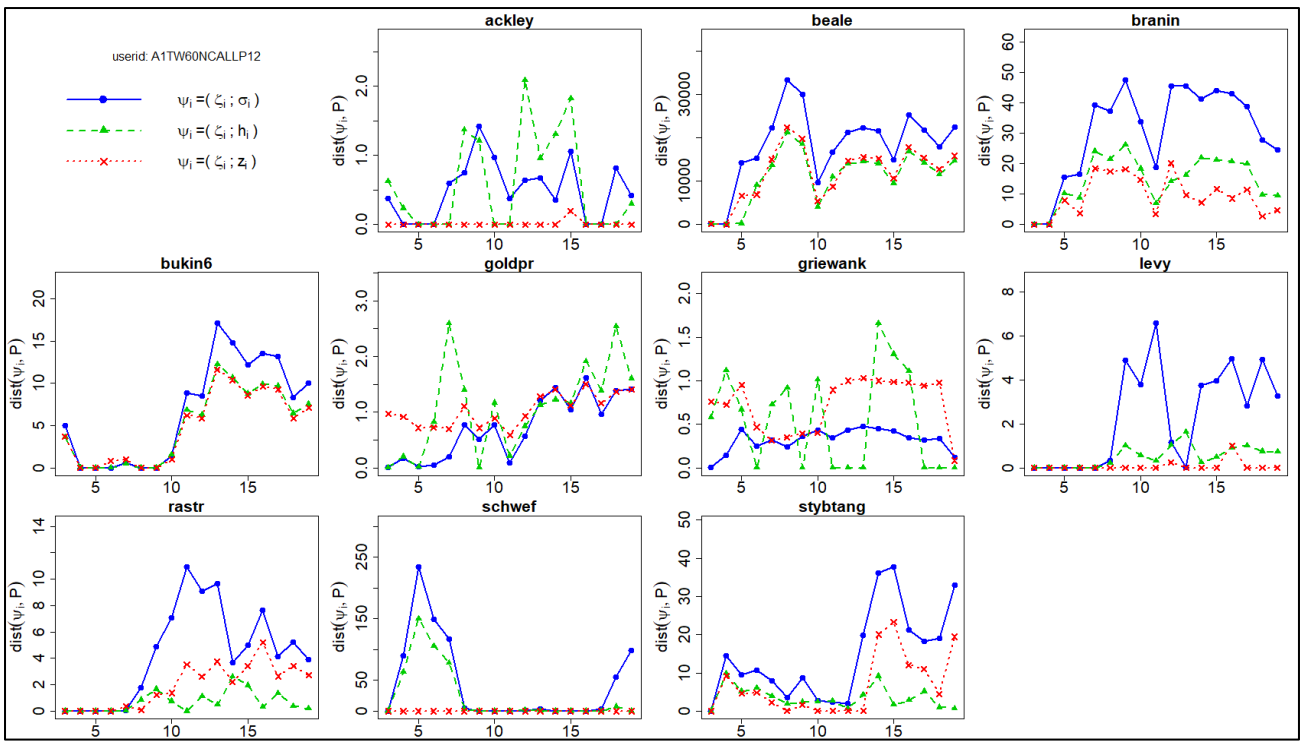

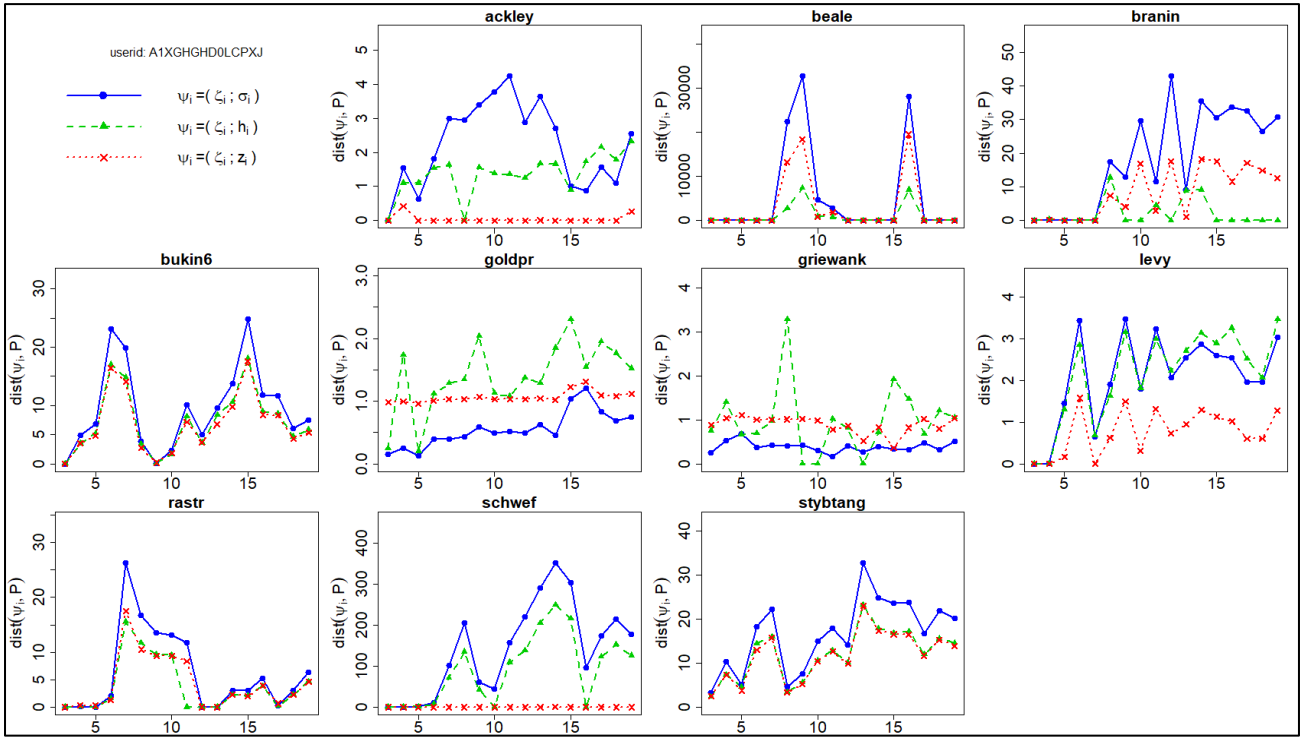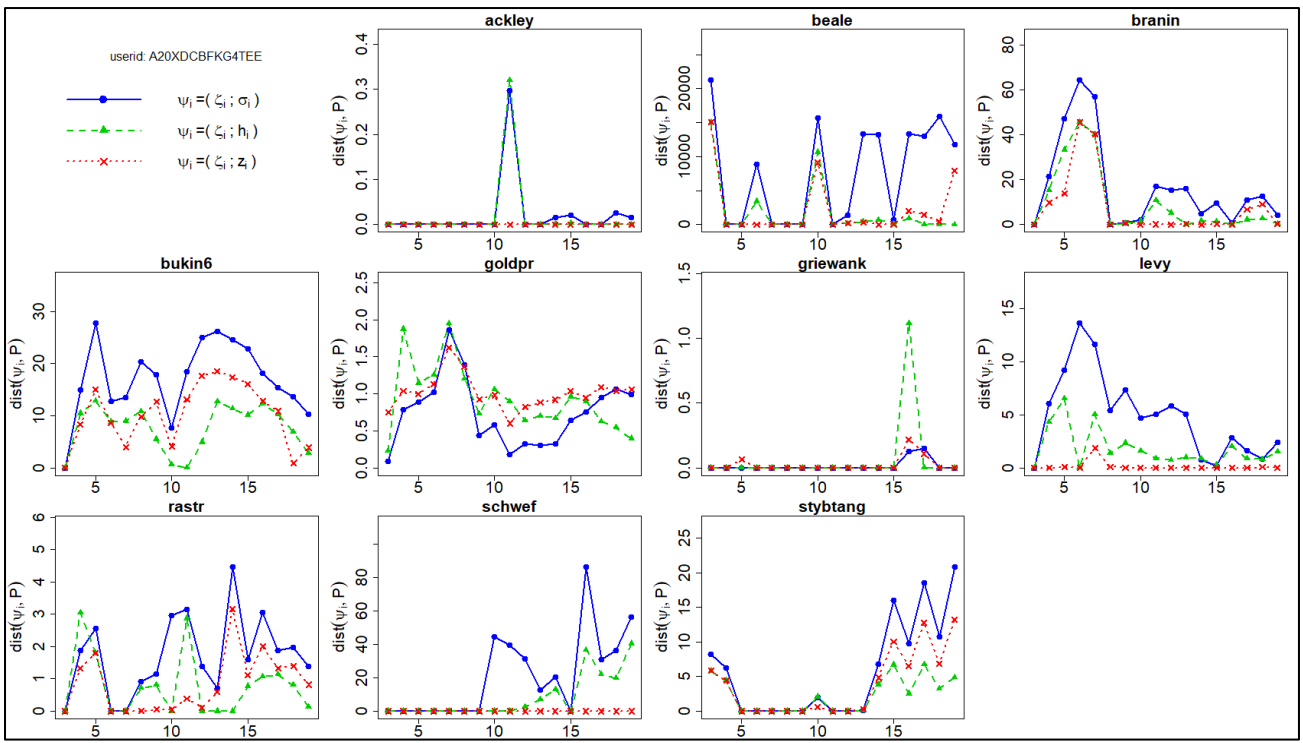

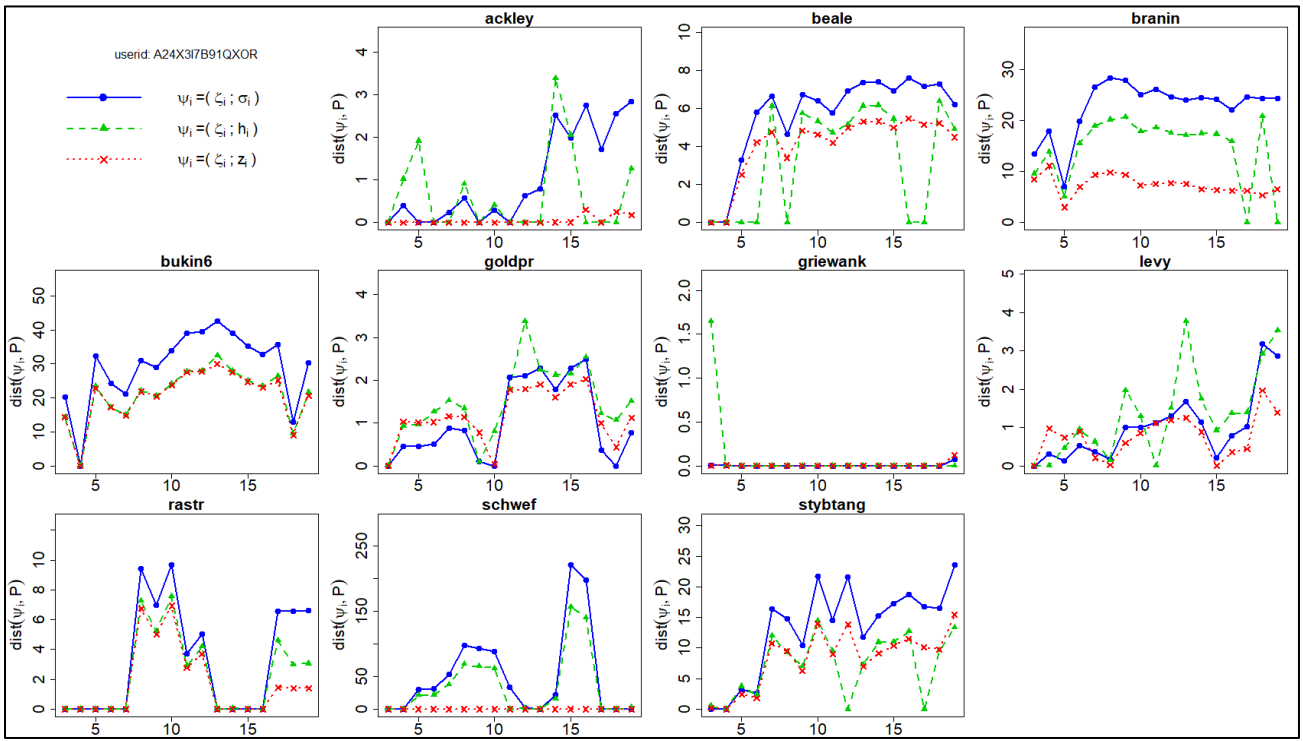

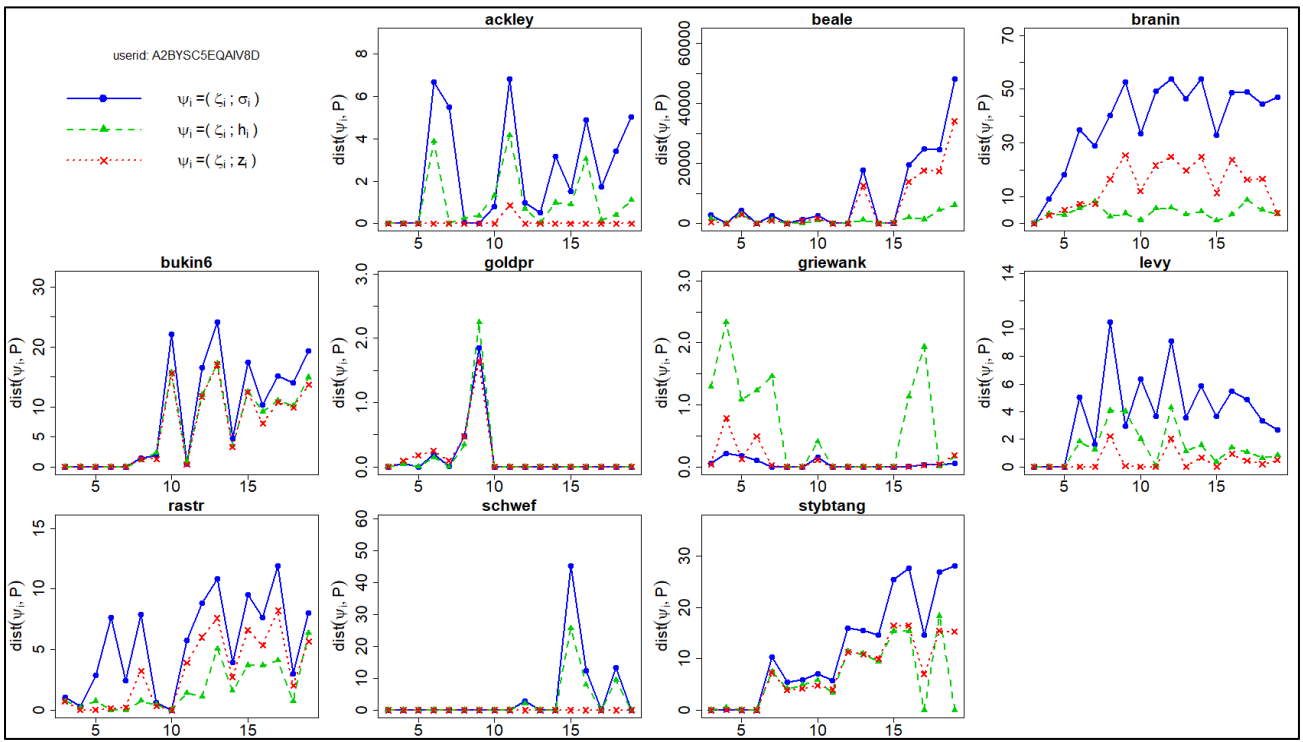

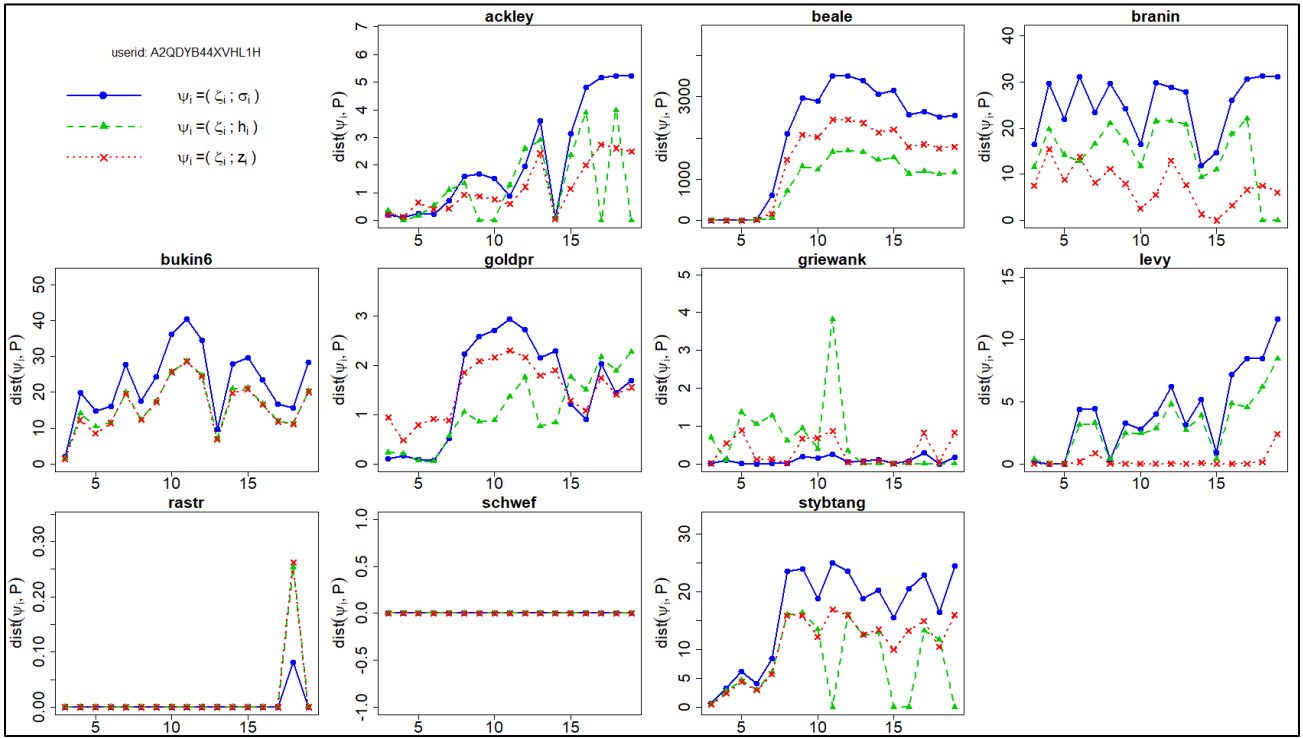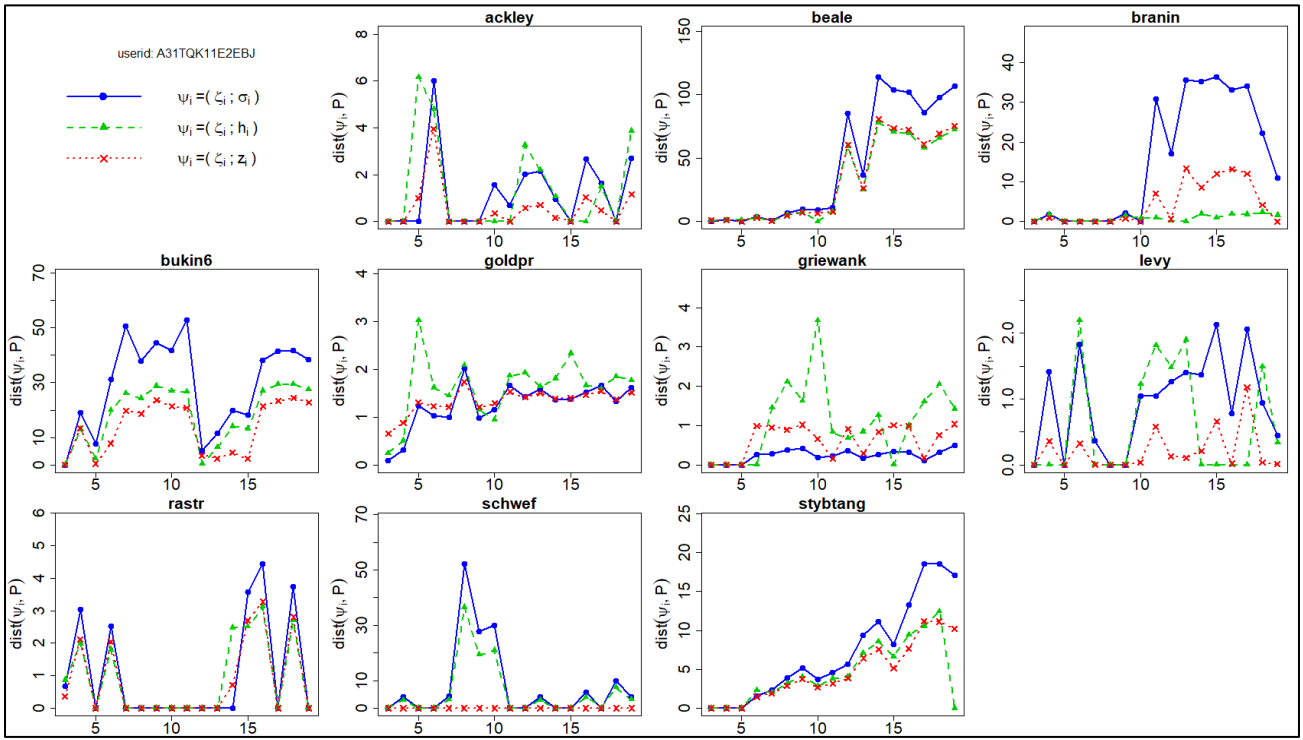

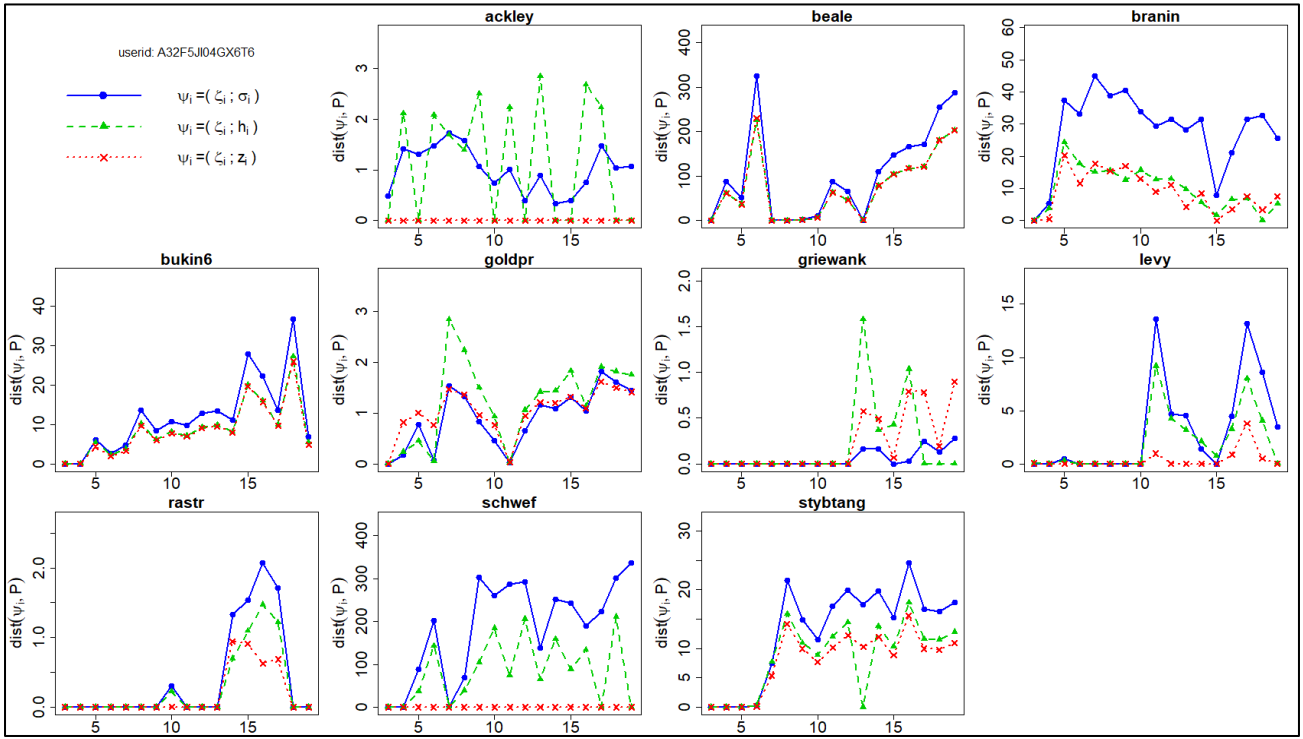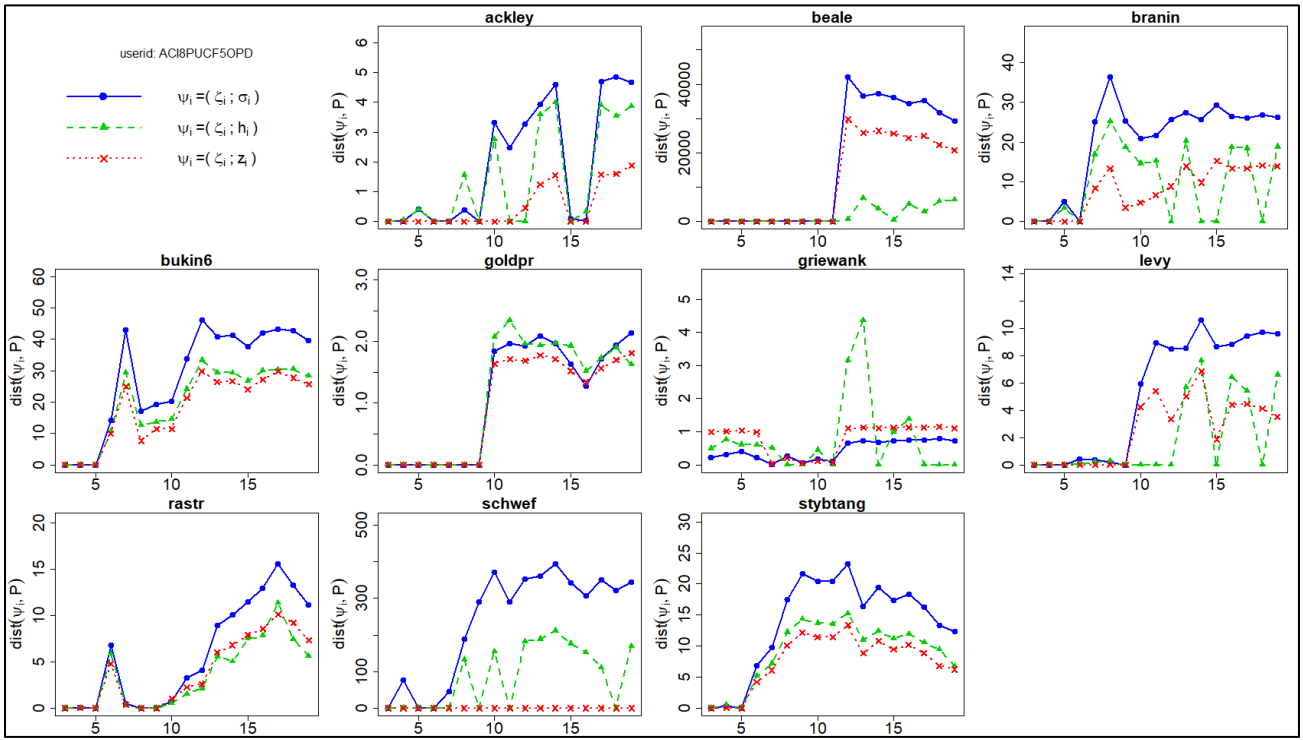

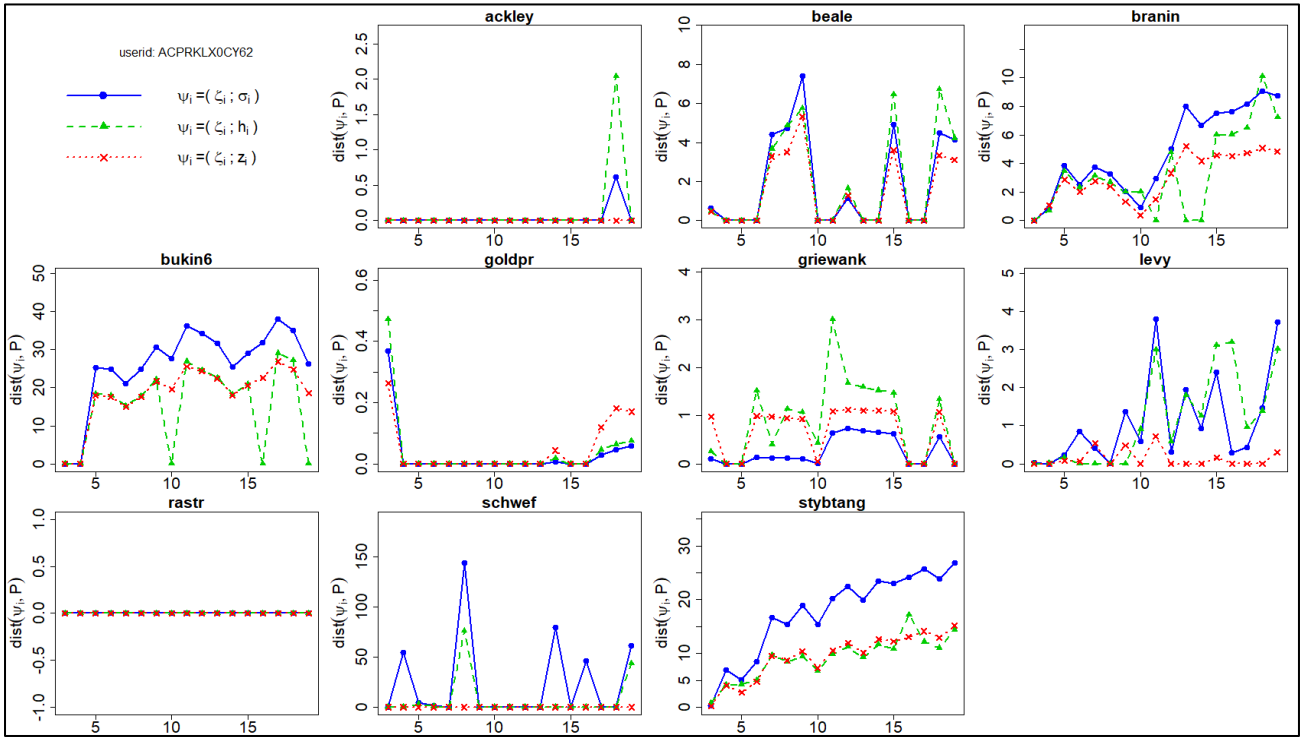
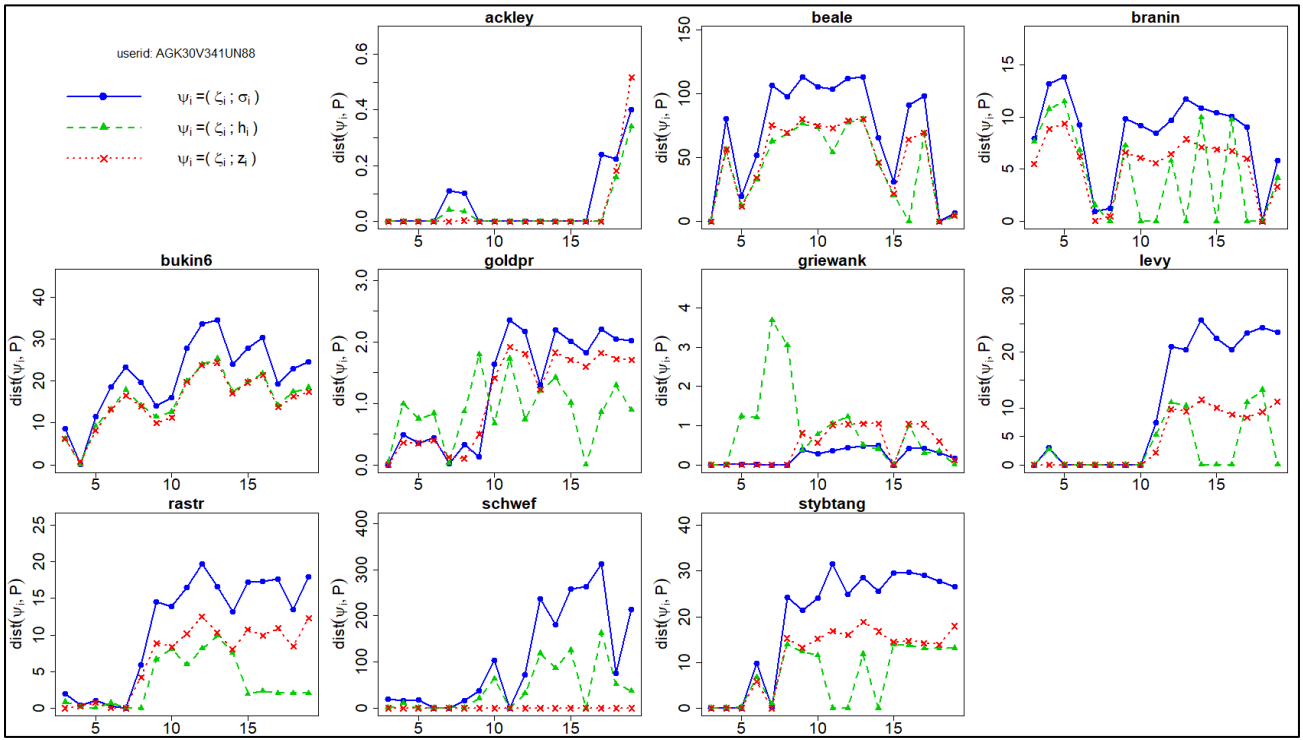